\documentclass[runningheads]{llncs}

\usepackage{eccv}

\usepackage{eccvabbrv}

\usepackage{graphicx}
\usepackage{booktabs}
\input{def_custom.tex}
\usepackage[accsupp]{axessibility}  
\usepackage{booktabs}
\usepackage{multirow}
\usepackage[table]{xcolor}
\usepackage{caption} 
\usepackage{adjustbox}
\usepackage{wrapfig}
\usepackage{epigraph} 
\usepackage{titletoc}

\usepackage{hyperref}
\definecolor{citeblue}{RGB}{0, 0, 200}
\hypersetup{
    colorlinks=true,
    citecolor=citeblue,
    linkcolor=citeblue,
    urlcolor=black
}

\usepackage{orcidlink}

\definecolor{azure}{rgb}{0.0, 0.5, 1.0}
\begin{document}

\title{See \& Sniff: Learning Visuo-Olfactory Representations} 

\titlerunning{See \& Sniff}

\author{Seongyu Kim$^{*}$\inst{1} \and
Seungwoo Lee$^{*}$\inst{1} \and
Hyeonggon Ryu\inst{2} \and
Joon Son Chung\inst{1} \and
Arda Senocak\inst{3}
}

\authorrunning{S. Kim et al.}
\definecolor{projectblue}{RGB}{0, 70, 180}
\institute{Korea Advanced Institute of Science and Technology, Korea \and
Hankuk University of Foreign Studies, Korea \and 
Ulsan National Institute of Science and Technology, Korea \\
\vspace{0.1em}
\href{https://mm.kaist.ac.kr/projects/SeeandSniff/}{\textcolor{projectblue}{\texttt{https://mm.kaist.ac.kr/projects/SeeandSniff}}}
}
\vspace{-0.6em}

\maketitle

\begin{center}
\centering
\vspace{-.15in}
\captionsetup{type=figure}
\includegraphics[width=\linewidth]{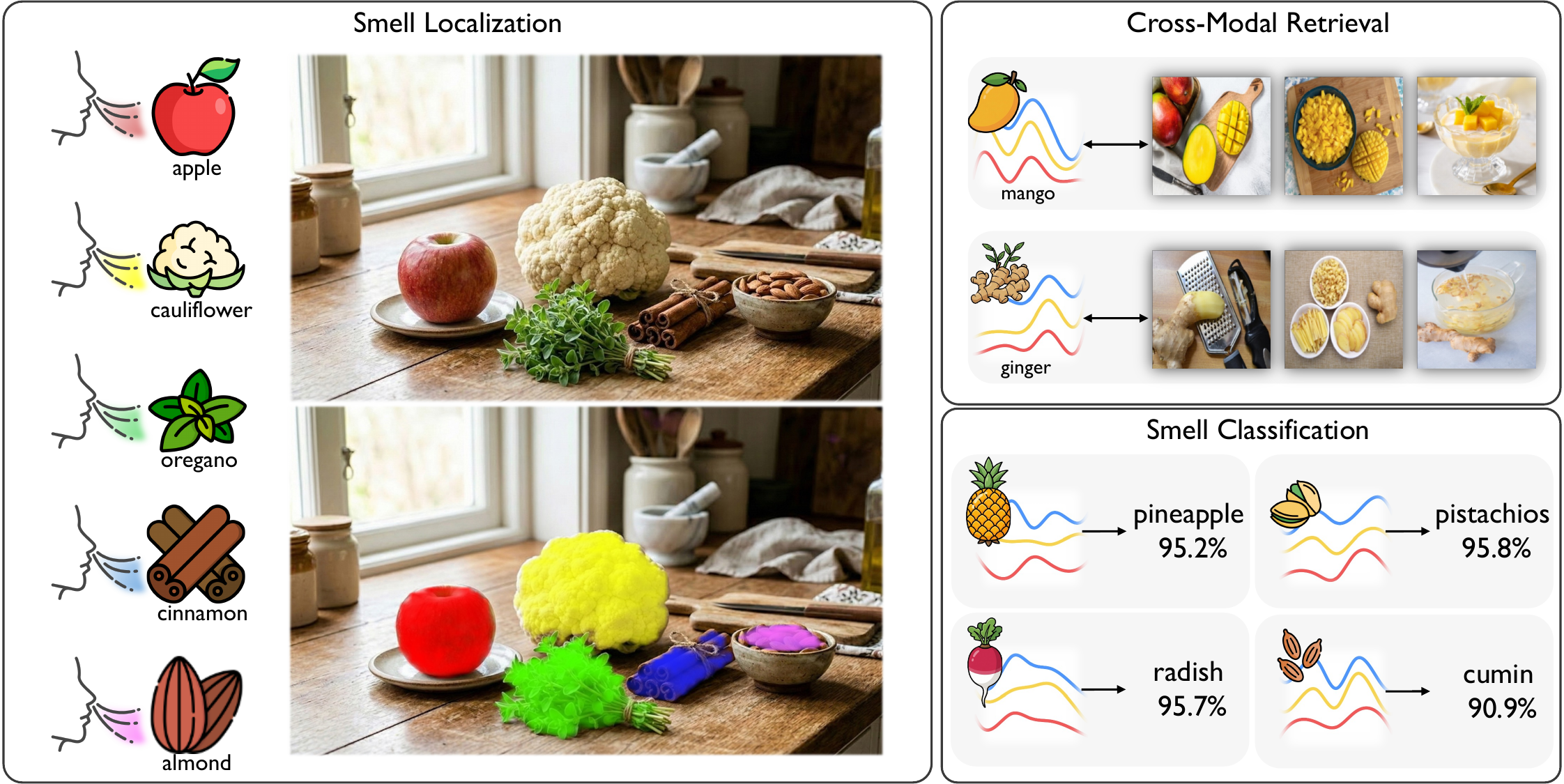}
\vspace{-2em}
\caption{\textbf{What can \textit{See \& Sniff} do?}
We show that \textit{See \& Sniff}, the framework that learns joint visuo-olfactory representations, can handle both unimodal and cross-modal tasks. (Left) Smell localization identifies the locations of smell sources within a visual scene based on input olfactory signals. (Right Top) Cross-modal retrieval demonstrates semantic alignment through bidirectional retrieval. (Right Bottom) Smell classification predicts ingredients from olfactory inputs.
}
\vspace{-7mm}
\label{fig:teaser}
\end{center}
\let\thefootnote\relax\footnotetext{$^{\star}$ Equal contribution.}
\begin{abstract}
While modern multimodal models integrate vision with language, audio, or touch, olfaction remains largely unexplored due to the lack of paired visuo-olfactory data. We introduce \textit{SmellNet-V}, a scalable visuo-olfactory dataset built on the insight that odor identity is largely invariant to visual transformations within a semantic category. This allows us to synthetically pair smell-only samples with semantically aligned in-the-wild web images, converting a unimodal olfactory dataset into a cross-modal benchmark without costly co-collection. Building on this dataset, we propose \textit{See \& Sniff}, a self-supervised framework that learns joint visuo–olfactory representations via dense local alignment and naturally produces smell saliency maps for spatial grounding of odor sources. We further introduce pixel-level smell localization task and a benchmark for evaluation. Our method surpasses smell-only baselines by 7\% in smell classification from smell alone and generalizes to cross-modal retrieval and smell localization, establishing visuo-olfactory learning as a new direction in multimodal perception. 
\vspace{-0.7em}
\end{abstract}
\section{Introduction}
\label{sec:intro}
\vspace{-1em}
\renewcommand{\epigraphflush}{flushright}
\renewcommand{\epigraphsize}{\small}
\setlength{\epigraphwidth}{0.65\textwidth}
{\scriptsize
\epigraph{\textit{Everything has its own odor, which is its soul—tree, flower, soil, rain, burning wood.}\hspace{30pt}\textit{Helen Keller}}
{}
}\vspace{-4mm}
\noindent Modern multimodal AI systems integrate vision with language~\cite{radford2021learning, jia2021scaling, li2022blip, li2023blip, zhai2023sigmoid}, audio~\cite{aytar2016soundnet, arandjelovic2017look, senocak2018learning, owens2018audio, harwath2018jointly, girdhar2023imagebind, gong2022contrastive}, and even touch~\cite{yang2022touch, yang2023generating, yang2024binding, fu2024a, lyu2024omnibind,kim2026seeing}, yet overlook another fundamental sensory modality: \textit{smell}. Despite its central role in biological perception, olfaction remains minimally explored in AI research, “\textit{a neglected treasure}”~\cite{keller1934neglected}. This gap comes primarily from practical constraints: unlike cameras or microphones, portable olfactory sensors have historically been scarce and confined to laboratory settings, and large-scale public datasets for data-driven learning have been lacking. Recent advances in portable chemical sensing now enable more accessible machine-readable olfactory signals, and the first large-scale smell dataset has very recently been released~\cite{feng2025smellnet}. However, it contains only olfactory measurements without paired visual data.

Why does visuo-olfactory integration matter? Human vision and olfaction are tightly intertwined sensory systems. When a person sees roasted coffee beans, they anticipate the characteristic bitter and roasted aroma even without smelling it. Conversely, the scent of coffee alone can evoke mental imagery of dark roasted beans or a steaming cup. These correspondences reveal a tight coupling between visual appearance and olfactory identity. Modeling this visuo-olfactory alignment computationally is central to our work.

Our objective in this work is to enable a self-supervised model to learn joint representations between visual and olfactory modalities, which fundamentally requires paired visuo-olfactory data. However, such datasets do not currently exist. We address this gap through \textit{a key insight}: olfactory identity is largely invariant to visual transformations, including changes in lighting, scale, or minor color variations. For example, intra-class variations such as \textit{a small red apple} and \textit{a large red apple} emit nearly identical volatile organic compounds (VOCs) and are perceptually categorized by humans under the same odor category, `\textit{apple}'. From a representation learning perspective, this suggests that visually diverse instances within a category should share a common semantic odor embedding. Leveraging this invariance, we construct a visuo-olfactory training dataset, \textit{SmellNet-V}, by synthetically pairing each odor instance from an existing smell-only dataset~\cite{feng2025smellnet} with semantically corresponding, in-the-wild open-world web images. This principled pairing strategy transforms a unimodal olfactory dataset into a scalable visuo-olfactory benchmark, enabling large-scale cross-modal alignment without the need for costly real-world paired data collection.

Multisensory perception extends beyond simple correspondence. Humans and other animals not only associate odors with visual identities, but also use olfactory cues to navigate and localize odor sources within complex environments~\cite{castellotti2025visual, raithel2021using}, \ie, \textit{smell localization}. 
To model this capability, we propose a framework that computes dense similarity maps between local visual and olfactory features, enabling fine-grained alignment. The resulting smell saliency maps highlight image regions expected to emit a given odor, while simultaneously learning holistic cross-modal correspondences. To evaluate smell localization, we introduce the first dataset with pixel-level segmentation aligned with the odor samples.

Overall, our proposed training data, \textit{SmellNet-V}, and self-supervised framework, \textit{See \& Sniff}, establish a new multimodal model for visuo-olfactory perception. The learned joint representations generalize across downstream tasks, including smell classification, cross-modal retrieval, and smell localization. In particular, for smell classification, our model surpasses the smell-only baseline by a significant margin, demonstrating that visuo-olfactory training enhances olfactory representation learning. Together, these contributions introduce new tasks and benchmarks for this emerging modality.

Our contributions are summarized as follows:
\vspace{-2mm}
\begin{itemize}
\item We construct a visuo-olfactory dataset, \textit{SmellNet-V}, by leveraging odor invariance within semantic categories to synthetically pair existing smell-only data with in-the-wild visual images.

\item We propose \emph{See \& Sniff}, a self-supervised multimodal framework that learns joint visuo–olfactory representations via dense local alignment.

\item We demonstrate that visuo-olfactory training enhances olfactory representation learning, surpassing the smell-only baseline by 7\% in smell classification.

\item We extend visuo-olfactory learning beyond classification and retrieval to spatial grounding by generating smell saliency maps and introducing the first pixel-level smell localization dataset for evaluation.

\end{itemize}
\section{Related Work}\label{sec:related}
Machine learning approaches to olfaction have primarily focused on predicting odor perception from chemical and molecular structure. Neural network–based models such as DeepNose~\cite{tran2019deepnose} demonstrated that artificial networks can learn structured embeddings of odorant space. Subsequent works extended this direction using graph neural networks~\cite{sanchez2019machine, lee2023principal, achebouche2022application}, multitask learning frameworks that capture shared representations across related odor categories~\cite{iwata2025interpretable}, and attention-based aggregation mechanisms built upon molecular representations derived from chemical foundation models~\cite{kang2026aromma}. Other studies have addressed specific challenges such as odor intensity prediction~\cite{fichtelmann2025machine}. Parallel efforts have explored artificial olfactory systems using sensor measurements. Electronic nose based frameworks, such as k-nearest neighbor–based scent classification method~\cite{mueller2019scent} and the eigengraph-based system~\cite{sung2024data}, focus on signal-level representation learning and odor classification. While effective, these approaches remain unimodal and are primarily developed in controlled laboratory environments, limiting their scalability to large-scale real-world settings.

More recently, SmellNet~\cite{feng2025smellnet} introduced the first large-scale real-world smell dataset collected with portable chemical sensors, along with ScentFormer, a transformer-based model for olfactory representation learning. While this represents a significant step toward scalable data-driven machine olfaction beyond laboratory environments, the framework remains unimodal and lacks paired visual information. Building upon SmellNet, we extend its smell-only data to the visual domain by synthetically pairing samples with semantically aligned in-the-wild images, leveraging odor invariance within semantic categories. Whereas ScentFormer focuses on unimodal odor modeling, our framework adopts a two-stream transformer with a dedicated local alignment module to learn joint visuo–olfactory representations and dense cross-modal correspondences.

A concurrent unpublished work by Ozguroglu \etal~\cite{ozguroglu2025new} also studies visuo-olfactory learning, focusing on large-scale naturally paired data collection using a handheld sensing device to demonstrate visual supervision for odor representation learning. In contrast, we extend a smell-only dataset via synthetic pairing based on odor invariance. Architecturally, while their method employs global contrastive alignment, ours incorporates dense local alignment, enabling both fine-grained spatial correspondence and global representation learning. Additionally, we introduce smell localization as a new downstream task for visuo-olfactory grounding. Nevertheless, both works highlight the timeliness and importance of advancing visuo-olfactory multimodal learning.
\vspace{-2mm}
\section{Methodology}\label{sec:method}
\begin{figure*}[t!]
    \centering
    \includegraphics[width=\linewidth]{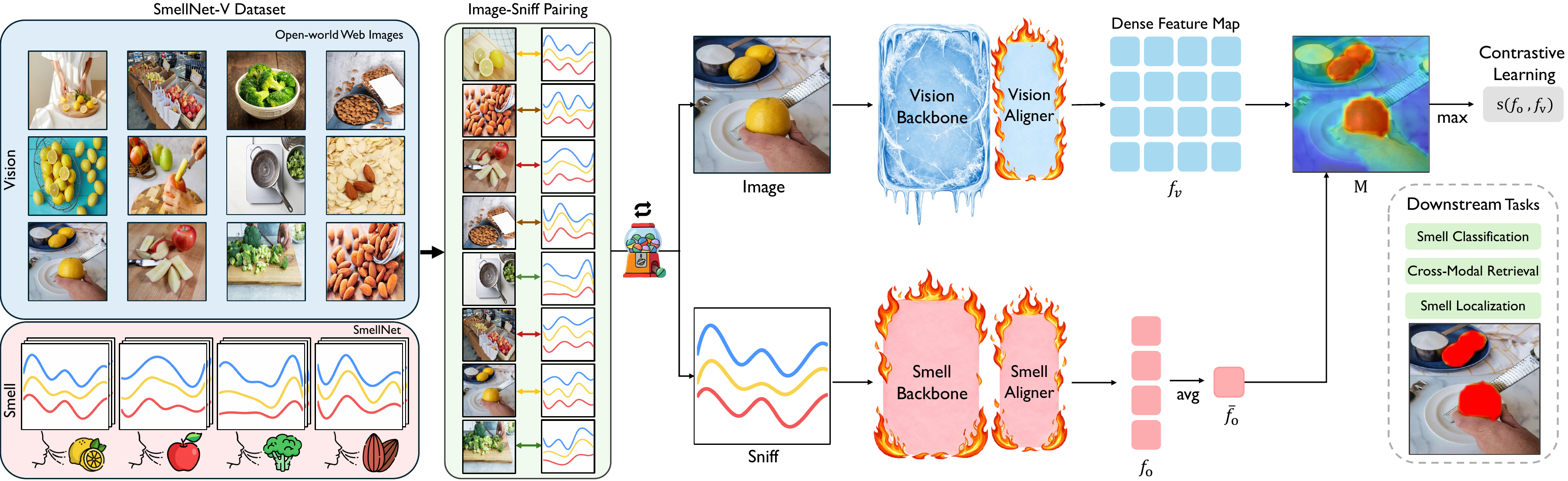}
    \caption{\textbf{Pipeline of \textit{See \& Sniff}.} Our framework expands smell-only data through semantic pairing with web images. Vision and smell encoders extract modality-specific features, which are aligned using a contrastive objective to learn joint visuo-olfactory representations.}
    \label{fig:pipeline}
    \vspace{-4mm}
\end{figure*}

\vspace{-2mm}
We aim to learn joint visuo-olfactory representations that generalize to downstream tasks including smell classification, cross-modal retrieval, and smell localization. To this end, we first construct a visuo-olfactory dataset by extending a smell-only dataset with semantically aligned in-the-wild web images. We then train a self-supervised framework that projects sniff and visual inputs into a shared embedding space and learns dense local cross-modal alignment via contrastive learning. An overview of the proposed framework is shown in~\Fref{fig:pipeline}.
\vspace{-5mm}
\subsection{Construction of \textit{SmellNet-V}}\label{sec:smellnetv}
\vspace{-1mm}
Learning joint visuo–olfactory representations requires paired multimodal training data. In the absence of such datasets, we construct a visuo-olfactory dataset, \textit{SmellNet-V}, by synthetically pairing odor samples in the smell-only dataset~\cite{feng2025smellnet} with semantically aligned in-the-wild web images. This pipeline produces multimodal data that approximate natural visuo-olfactory correspondences while increasing data granularity and enabling finer cross-modal alignment.

\noindent\textbf{SmellNet Overview.} SmellNet~\cite{feng2025smellnet} is a large-scale dataset capturing real-world olfactory signals from 50 food and natural ingredients, grouped into nuts, spices, herbs, fruits, and vegetables, collected using portable multi-channel gas sensors. For each ingredient, 10 minutes of data were recorded per six sessions on different days, resulting in approximately 180,000 time-series samples at 1~Hz. Although 12 sensor channels are available, prior work primarily utilizes 6 chemically relevant components: $\text{NO}_2$, $\text{C}_2\text{H}_5\text{OH}$, VOC, CO, Alcohol and LPG. Importantly, SmellNet captures ingredients in their normal, non-degraded states.

\noindent\textbf{Sniffing Units.} In extending SmellNet to the visual domain, we draw inspiration from biological olfaction, where perception occurs through brief sniffing episodes rather than prolonged exposure~\cite{mainland2006sniff, wachowiak2011all}. Accordingly, we segment longer olfactory recordings into fixed-length temporal windows of size $W$ with stride $s$, analogous to individual `\textit{sniffs}', and treat each sniff as a discrete training sample. This increases data granularity and aligns the training paradigm with natural olfactory sampling behavior.

\noindent\textbf{Image Collection.} To obtain diverse visual samples, we retrieve web images using context-rich queries for each ingredient. Rather than relying solely on class names, we use a large language model (LLM)~\cite{singh2025openai} to generate descriptive phrases that capture varied real-world scenarios associated with each SmellNet ingredient. These queries are then used to collect visually diverse images (prompting details are provided in the suppl. material). For instance, for the ingredient `\textit{apple}', the LLM generates phrases such as `\textit{whole apple on wooden table}', `\textit{apples in grocery produce section}', and `\textit{apple hanging on tree}'. By covering diverse environments and object configurations, the resulting image set captures broad visual variability within each ingredient category.

\noindent\textbf{Image Filtering.} After image collection, we apply a three-stage filtering pipeline. 
(1) \textit{Prompt-based verification}: We use CLIP~\cite{radford2021learning} with structured positive--negative prompt pairs to enforce (i) category consistency, (ii) photorealism (excluding drawings or illustrations), and (iii) valid object state (excluding spoiled or degraded instances). Images are retained only if their similarity to the positive prompt exceeds that of the corresponding negative prompts in all three tests. The exact prompt formulations are provided in the suppl. material. 
(2) \textit{Quality filtering}: Images containing excessive watermarks (more than five detections using an off-the-shelf watermark detector~\cite{fancyfeast2025watermark}) are removed. 
(3) \textit{Human refinement}: Finally, annotators perform lightweight manual verification to discard any remaining irrelevant samples.

\noindent\textbf{Image-Sniff Pairing.} After image collection and filtering, we construct visuo-olfactory training pairs by randomly matching each sniff unit with an image belonging to the same ingredient category. This category-level pairing yields discrete visuo–olfactory samples for training and forms the \textit{SmellNet-V}.

\noindent\textbf{Design Rationale of \textit{SmellNet-V}.}
Our extension relies on the observation that odor identity is largely invariant to common visual transformations, including variations in lighting, scale, and appearance differences. For example, \textit{a small red apple} and \textit{a large red apple} emit nearly identical VOC signatures and are perceived under the same odor category, \textit{apple}. This invariance allows us to pair any sniff unit within a ingredient category with visually diverse web images that share the same semantic identity. Moreover, because SmellNet captures ingredients in their normal, non-degraded states, visually matched web images in corresponding normal conditions are readily available, making such pairing physically consistent and semantically reliable. Thus, the resulting \textit{SmellNet-V} approximates natural visuo-olfactory correspondences.
\vspace{-2mm}

\subsection{Model and Training Objective}

\noindent\textbf{Contrastive Learning} aims to learn aligned representations by attracting positive pairs while repelling negative pairs. In the visuo-olfactory setting, let $E_o$ and $E_v$ denote the olfactory and visual encoders, respectively. For a sniff unit $o_i$, we obtain its feature ${f}_{o_i}{=}E_o(o_i)$, and for the corresponding visual sample $v_i$, we compute ${f}_{v_i}{=}E_v(v_i)$. The remaining visual features ${f}_{v_j}$ for $j \neq i$, drawn from the dataset $\calD = \{(v_i,o_i)\}_{i=1}^N$, serve as negatives. The objective $\calL$ is defined as:

\begin{equation}
\label{eq:ov_contrastive}
\calL = -\log \frac{\exp(s(f_{o_i}, f_{v_i})/\tau)}{\sum_j \exp(s(f_{o_i}, f_{v_j})/\tau)},
\end{equation}

\noindent where $s(\cdot,\cdot)$ denotes a cross-modal similarity function and $\tau$ is a temperature parameter~\cite{wu2018unsupervised}. Following prior studies~\cite{senocak2023sound,radford2021learning,elizalde2022clap,jia2021scaling, girdhar2023imagebind,yang2024binding}, we employ the loss in a symmetric manner across modalities.

\noindent\textbf{Vision and Olfactory Encoders.}
Given an image $v_i$ and its corresponding sniff unit $o_i$, the respective encoders extract modality-specific representations. Each encoder is composed of a backbone network followed by a lightweight aligner module. Through this pipeline, the raw inputs are projected into a shared representation space, producing a visual feature map $f_{v_{i}} \in \mathbb{R}^{C \times H \times W}$ and a sniff feature $f_{o_{i}} \in \mathbb{R}^{T \times C}$. Here, $H$ and $W$ denote the height and width of the feature map, $T$ denotes the temporal sequence length of a sniff unit, and $C$ indicates the channel dimension of the common embedding space.

\noindent\textbf{Similarity Function.} To enable fine-grained visuo-olfactory alignment, we adopt a similarity function that accounts for the characteristics of the task. Since our goal is to learn joint visuo-olfactory representations via dense local alignment, we first aggregate the sniff feature into a average pooled vector
$\bar{f}_{o_{i}} {=} \frac{1}{T} \sum_{t=1}^T f_{o_{i}}[t]$, where ${f}_{o_{i}}[t] \in \mathbb{R}^{C}$ refers to the one-dimensional vector at the time step $t$ of $f_{o_{i}} \in \mathbb{R}^{T\times C}$.
We then compute a similarity map $M \in \mathbb{R}^{H \times W}$ by measuring the similarity between the spatial visual feature map and the aggregated sniff feature: 
$M[h,w] = \bar{f}_{o_{i}} \cdot f_{v_{i}}[h, w]$,
where $f_{v_{i}}[h, w] \in \mathbb{R}^{C}$ denotes the one-dimensional feature vector at spatial location $[h, w]$, and $\cdot$ indicates the inner product. We max-pool the similarity map to get the final similarity score
$s(f_{o_{i}},f_{v_{i}}) {=} \max(M)$.

\vspace{-2mm}
\subsection{Implementation Details}

\noindent\textbf{Model Architecture.} Our framework consists of a visual encoder $E_v(\cdot)$ and an olfactory encoder $E_o(\cdot)$. For the visual branch, we adopt DINOv3-Small~\cite{simeoni2025dinov3} pretrained weights and keep the backbone frozen during training. For the olfactory branch, we follow the ScentFormer from SmellNet~\cite{feng2025smellnet}, implemented as a 4-layer transformer~\cite{vaswani2023attentionneed} with 8 attention heads, trained from scratch. To project modality-specific features into a shared embedding space, we append lightweight aligners to each backbone. For the vision encoder, we use a channel-wise LayerNorm~\cite{ba2016layernormalization} followed by a $1\times1$ convolution, which largely preserves the pretrained representation while adapting it to our objective. For the smell encoder, we employ a two-layer MLP with a residual connection. Both aligners are trained jointly under the proposed objective.

\noindent\textbf{Training Pairs.} The initial visuo-olfactory pairs in \textit{SmellNet-V} are constructed via random ingredient-level matching between sniff units and web images as explained in~\Sref{sec:smellnetv}. As the visuo-olfactory pairs are synthetically constructed rather than co-collected, we re-sample image–sniff pairs at each epoch to increase pairing diversity during training. Specifically, while the set of sniff units remains fixed, images within each ingredient are cyclically shifted. This strategy exposes each sniff unit to multiple visual instances across epochs, encouraging robust ingredient-level alignment rather than memorization of fixed synthetic pairs.

\noindent\textbf{Training Setup.} Our model takes a $224\times224$ image and a single sniff unit as input. For olfaction, we compute a first-order temporal difference with lag $p$, $\Delta x_{t}=x_{t}-x_{t-p}$, to emphasize relative changes in qualitative sensor outputs. We then extract windows of size $(W, N_s)$ with stride $W/2$. The number of sensor channels $N_s$ is 6, and we use $p{=}25$ and $W{=}50$ by default, unless specified otherwise. Training is performed on a NVIDIA RTX A5000 GPU with a batch size of 64. See suppl. for details. 
\section{Experiments}\label{sec:experiments}
We evaluate our visuo-olfactory representations on \textit{smell classification}, \textit{cross-modal retrieval}, and \textit{smell localization}. This section describes the datasets, baselines, and results.
\subsection{Datasets}\label{ssec:datasets}

\noindent\textbf{Training dataset.}
Our visuo-olfactory model is trained on \textit{SmellNet-V}. With the default setting, \textit{SmellNet-V} yields 5,411 sniff--image pairs for training.

\noindent\textbf{Testing datasets.}
We evaluate our method on task-specific test sets as follows:
\begin{itemize}
\item \textbf{SmellNet-Test}~\cite{feng2025smellnet}: The official test split of SmellNet, used for smell classification from smell alone. With the default setting, it yields 1,083 sniffs.

\item \textbf{\textit{SmellNet-V}-Test:} A visuo-olfactory extension of the SmellNet test set constructed following our procedure in~\Sref{sec:smellnetv}. It is used for cross-modal retrieval evaluation. The dataset contains the same number of smell samples as SmellNet-Test.

\item \textbf{\textit{SmellNet-V}-Source:} A newly constructed benchmark for smell source localization. Built upon \textit{SmellNet-V}-Test, it includes manually annotated ingredient segmentation masks derived from the ground-truth ingredient categories. We use an interactive annotation tool~\cite{cvat} powered by SAM~\cite{kirillov2023segment} to produce segmentation masks. Annotators provide sparse point prompts via simple mouse clicks to annotate regions according to predefined ingredient categories. Example annotations are in the second column of~\Fref{fig:localization}.
\end{itemize}
\vspace{-4mm}
\subsection{Baselines}
We compare our model with representative baselines, grouped by evaluation purpose, and specify the baselines used for each task.

\noindent\textbf{Olfactory-Only \vs Visuo-Olfactory.} To assess the impact of visual supervision on olfactory representation learning, we use ScentFormer from SmellNet as a unimodal baseline, evaluated only on ingredient smell classification task.

\noindent\textbf{Global \vs Local Alignment.} Self-supervised multimodal contrastive learning typically aligns modalities using inner products between global representations, such as pooled features or class tokens. While effective for tasks requiring holistic understanding (\eg, linear probing or retrieval), global alignment may be insufficient for spatial localization. In contrast, dense local alignment based on similarity maps may enable both global correspondence and fine-grained grounding. The following baselines are included to compare global and local alignment strategies: 
(1) \textit{See \& Sniff}: Our proposed model trained with the dense local alignment objective, (2) \textit{Ours-Local}: Local alignment model without aligners for each modality encoder, (3) \textit{Ours–Global}: A variant of \textit{Ours-Local} using a CLS-token based global alignment objective, (4) \textit{Global-CLIP}: A baseline combining a frozen CLIP-Large~\cite{radford2021learning} pre-trained visual encoder with our olfactory encoder, optimized using a CLIP-style global contrastive objective. The global baselines also approximate the recent work~\cite{raithel2021using}. These baselines are used for all tasks.\\

\noindent Below baselines are only used for smell localization:

\noindent\textbf{Visual Bias Analysis.} To examine potential visual biases in the localization benchmark, we include several vision-only baselines that do not rely on olfactory input. (1) \textit{Full Square} and (2) \textit{Full Circle}: fixed binary masks (a $224 \times 224$ square or a circle with diameter 224) applied uniformly without any visual or smell understanding. (3) \textit{DINOv3 Attention Map}: a vision-only baseline derived from attention maps of a pre-trained DINOv3 model, capturing generic objectness cues without smell information.

\noindent\textbf{Cascaded Approach.} We consider a two-stage pipeline, SmellNet + SAM3, that first predicts smell categories using SmellNet and subsequently performs text-conditioned segmentation (\eg, SAM3~\cite{carion2025sam}) for localization. This cascaded approach serves as a baseline to compare against our \textit{See \& Sniff} model.

\noindent\textbf{Upper Bound Baseline.}
We also include an upper-bound reference based on the cascaded approach. Here, ground-truth ingredient category labels are fed to SAM3 to get segmentation masks. This baseline estimates the maximum achievable localization when the smell classification is assumed to be perfect.

\vspace{-2mm}
\subsection{Main Results}
\noindent\textbf{-- Smell Classification Task}

\begin{figure*}[t]
  \centering
  \begin{minipage}[t]{0.56\textwidth}
    \vspace{3pt}
    \centering
    \small
    \adjustbox{max width=\linewidth}{%
      \begin{tabular}{lcccc}
        \toprule
        \multirow{2}{*}{\textbf{Model}} &
        \multicolumn{2}{c}{\textbf{$p=25, W=50$}} &
        \multicolumn{2}{c}{\textbf{$p=25, W=100$}} \\
        \cmidrule(lr){2-3} \cmidrule(lr){4-5}
        & \textbf{Acc.} & \textbf{F1} & \textbf{Acc.} & \textbf{F1} \\
        \midrule
        SmellNet (LSTM)        & 50.6  & 48.8  & 57.9  & 56.0  \\
        SmellNet (Transformer) & 50.6  & 49.5  & 56.1  & 55.5  \\
        Global-CLIP            & 53.19 & 52.22 & 61.55 & 60.01 \\
        Ours-Global            & 53.74 & 52.64 & 59.36 & 57.96 \\
        Ours-Local             & \underline{54.94} & \underline{53.78} & \underline{62.75} & \underline{62.31} \\
        \rowcolor{azure!10} 
        See \& Sniff  & \textbf{57.71} & \textbf{56.68} & \textbf{63.75} & \textbf{62.66} \\
        \bottomrule
      \end{tabular}
    }

    \captionof{table}{\textbf{Smell Classification.}}
    \label{tab:odor_cls}
  \end{minipage}
  \hfill
  \begin{minipage}[t]{0.428\textwidth}
    \vspace{1.5pt}
    \centering
    \includegraphics[width=\linewidth]{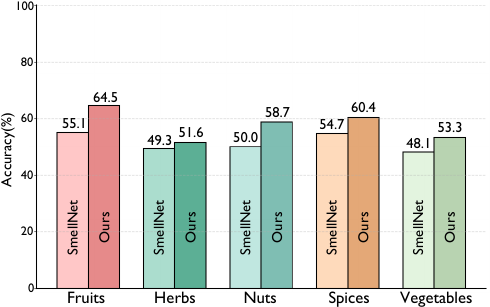}
    \vspace{-1.85em} 
    \captionof{figure}{\textbf{Family-wise Comparison.}}
    \label{fig:cat_lp}
  \end{minipage}
  \vspace{-6mm}
\end{figure*}

\noindent We evaluate the quality of learned olfactory representations on ingredient classification across the 50 categories defined in SmellNet. After self-supervised training of \textit{See \& Sniff} on \textit{SmellNet-V}, we freeze the olfactory encoder and train a linear probe using unimodal olfactory signals with the labels from SmellNet. Evaluation is conducted on the smell-only SmellNet-Test split. Results are in~\Tref{tab:odor_cls}. \textbf{Key findings} are as follows:

\noindent (1) \textit{Visual supervision via SmellNet-V and See \& Sniff learns stronger olfactory representations.} Across all settings and variants, \textit{See \& Sniff} trained on \textit{SmellNet-V} surpasses smell-only SmellNet baselines under smell-only inference, demonstrating that vision provides effective supervision for olfactory representation learning and validating our synthetic visuo-olfactory pairing strategy.

\noindent (2) \textit{The gains are robust across sniff configurations.} We report results under SmellNet’s original configurations ($p{=}25, W{=}50$ and $p{=}25, W{=}100$) for fair comparison. Under all settings, our model consistently outperforms SmellNet baselines, showing that performance gains are not tied to a specific parameter choice.

\noindent (3) \textit{Dense local alignment with aligner yields the best performance.} While global alignment provides competitive performance for holistic understanding, dense local alignment consistently achieves superior results. Furthermore, incorporating the lightweight aligner yields additional gains, supporting our architectural design for effective unimodal and cross-modal representation learning.

\noindent\textbf{Additional Analysis.} We further analyze smell classification across the five major ingredient families in SmellNet, including fruits, herbs, nuts, spices, and vegetables. Specifically, we report averaged ingredient-level classification results within each family to examine whether the observed gains are consistent across families rather than driven by a specific group. As shown in~\Fref{fig:cat_lp}, our model consistently improves performance across all five families compared to the smell-only baseline. Notable gains are observed for fruits, nuts, vegetables, and spices, with a modest but consistent improvement for herbs. The larger improvements in fruits, nuts, and vegetables can be attributed to their distinctive object-level visual structures, such as clear shape, color, and geometry, which provide informative cross-modal supervision when paired with olfactory signals. In contrast, herbs often exhibit fine-grained leaf textures and visually similar appearances across ingredients, limiting the additional discriminative cues available from visual supervision. Nevertheless, the consistent improvement across all ingredient families suggests that visuo-olfactory pairing enhances olfactory representation learning in a broad and family-agnostic manner.

\noindent\textbf{-- Cross-Modal Retrieval Task}

\noindent We evaluate cross-modal retrieval to measure the alignment between visual and olfactory embeddings. Given a sniff, the task is to retrieve the image of the same ingredient, and vice versa. Retrieval results are obtained by ranking the cosine similarity between the query embedding and embeddings from the other modality. Results are reported on the \textit{SmellNet-V} test set using Recall@K as the metric and are shown in~\Tref{tab:retrieval}.

\begin{table}[t]
\centering
\small
\setlength{\tabcolsep}{7pt}
\begin{tabular}{lccc|ccc}
\toprule
\multirow{2}{*}{\textbf{Model}} &
\multicolumn{3}{c}{\textbf{Smell $\rightarrow$ Vision}} &
\multicolumn{3}{c}{\textbf{Vision $\rightarrow$ Smell}} \\
\cmidrule(lr){2-4}\cmidrule(lr){5-7}
& \textbf{R@1} & \textbf{R@5} & \textbf{R@10} & \textbf{R@1} & \textbf{R@5} & \textbf{R@10} \\
\midrule
Global-CLIP            & 53.28 & 55.96 & 58.82 & \underline{56.90} & \textbf{87.40} & \underline{91.40} \\
Ours-Global            & 50.97 & \textbf{62.60} & \textbf{67.50} & 52.80 & 78.30 & 85.30 \\
Ours-Local             & \underline{53.65} & 58.45 & 61.50 & 55.40 & 82.90 & 88.90 \\
\rowcolor{azure!10}
See \& Sniff  & \textbf{56.14} & \underline{60.94} & \underline{63.90} & \textbf{63.20} & \textbf{87.40} & \textbf{91.90} \\
\bottomrule
\end{tabular}
\vspace{0.5em}
\caption{\textbf{Cross-modal Retrieval.}}
\label{tab:retrieval}
\vspace{-8mm}
\end{table}
\begin{figure*}[t!]
    \centering
    \includegraphics[width=0.98\linewidth]{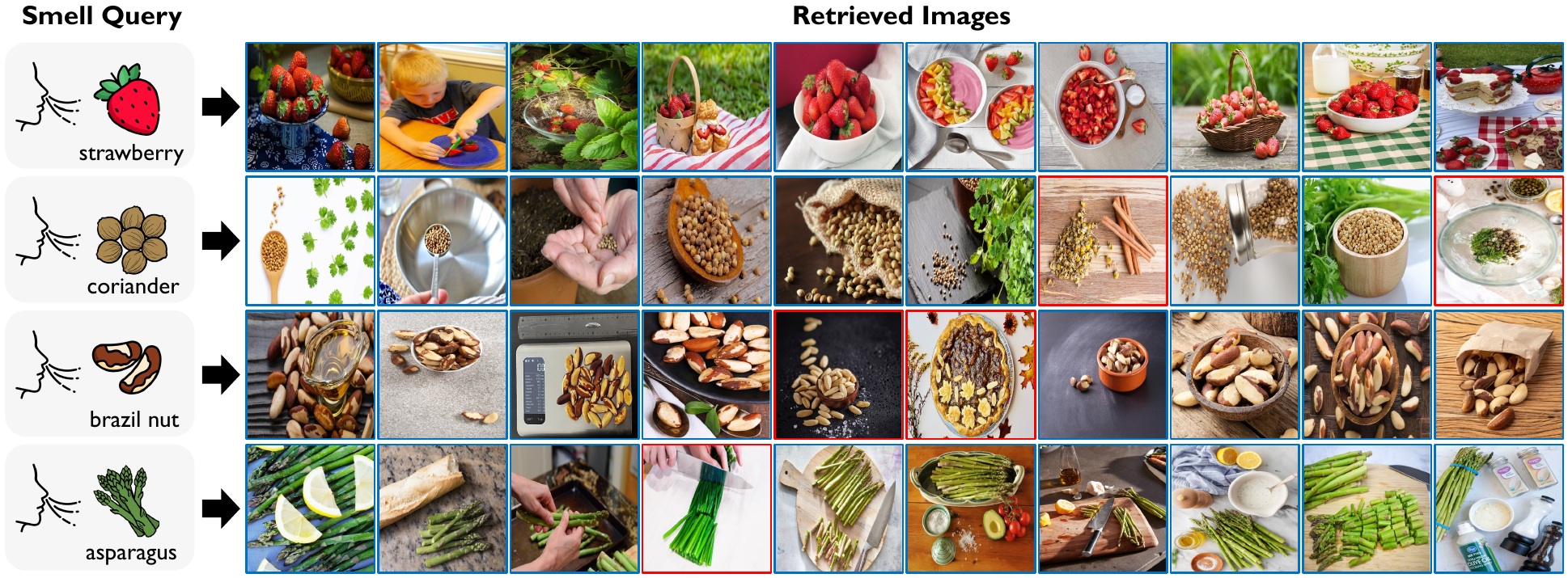}
    \vspace{-2mm}
    \caption{\textbf{Smell→Vision Retrieval}. The top--10 images retrieved by given smell queries are shown. Blue borders indicate correct matches, and red borders indicate mismatches. The failure cases are visually similar to the queried ingredients, indicating fine-grained visual ambiguity rather than random mismatch.
    }
    \label{fig:retrieval}
    \vspace{-6mm}
\end{figure*}

\noindent\textbf{Quantitative Results.} The retrieval results demonstrate that \textit{SmellNet-V} provides a viable foundation for visuo-olfactory training, as all variants exhibit cross-modal understanding, and that \textit{See \& Sniff} effectively learns cross-modal correspondences. It achieves the best overall performance in 4 of 6 metrics. As expected, global alignment performs strongly in retrieval, particularly on metrics such as R@5 and R@10 for Smell→Vision; however, our final model consistently outperforms it in the more challenging R@1 setting for both directions, indicating stronger fine-grained matching between visual and olfactory embeddings. Incorporating the aligner further improves performance, suggesting a more stable shared representation space that benefits cross-modal retrieval. Together with the improvements observed in smell classification and cross-modal retrieval, these results validate both our dataset construction strategy and alignment design.

\noindent\textbf{Qualitative Results.} The results are in~\Fref{fig:retrieval}. The model retrieves semantically consistent images conditioned on the input smell. For instance, the strawberry smell retrieves images of strawberries across diverse contexts, including sliced fruits, toping on the cake, and natural scenes. Similarly, the asparagus smell retrieves relevant asparagus images. A failure case (marked in red) retrieves chives, which are visually similar due to their elongated green structure. This error reflects fine-grained visual ambiguity rather than random mismatch.

\begin{wraptable}{r}{0.48\columnwidth}
\vspace{-2.2em}
\centering
\small
\setlength{\tabcolsep}{5pt}
\begin{tabular}{lccc}
\toprule
\textbf{Method} & \textbf{R@1} & \textbf{R@5} & \textbf{R@10} \\
\midrule
SmellNet & 52.54 & 76.18 & 85.50 \\
\rowcolor{azure!10}
See \& Sniff     & \textbf{69.44} & \textbf{85.04} & \textbf{89.47} \\
\bottomrule
\end{tabular}
\vspace{-0.9em}
\caption{\textbf{Smell→Smell Retrieval.}}
\label{tab:smell_uni_retrieval}
\vspace{-1.5em}
\end{wraptable}

\noindent\textbf{Smell→Smell Retrieval.} 
We further evaluate Smell→Smell retrieval (in~\Tref{tab:smell_uni_retrieval}) to assess the intrinsic structure of the learned olfactory embeddings. Compared to the SmellNet baseline, our model achieves superior retrieval performance, indicating that visuo-olfactory training enhances the the semantic structure of the smell embedding space. Since retrieval relies solely on smell features at inference time, these gains suggest that visual supervision also regularizes and strengthens intra-modal representations rather than simply enabling cross-modal matching. 

\begin{figure*}[t!]
    \centering
    \includegraphics[width=0.99\linewidth]{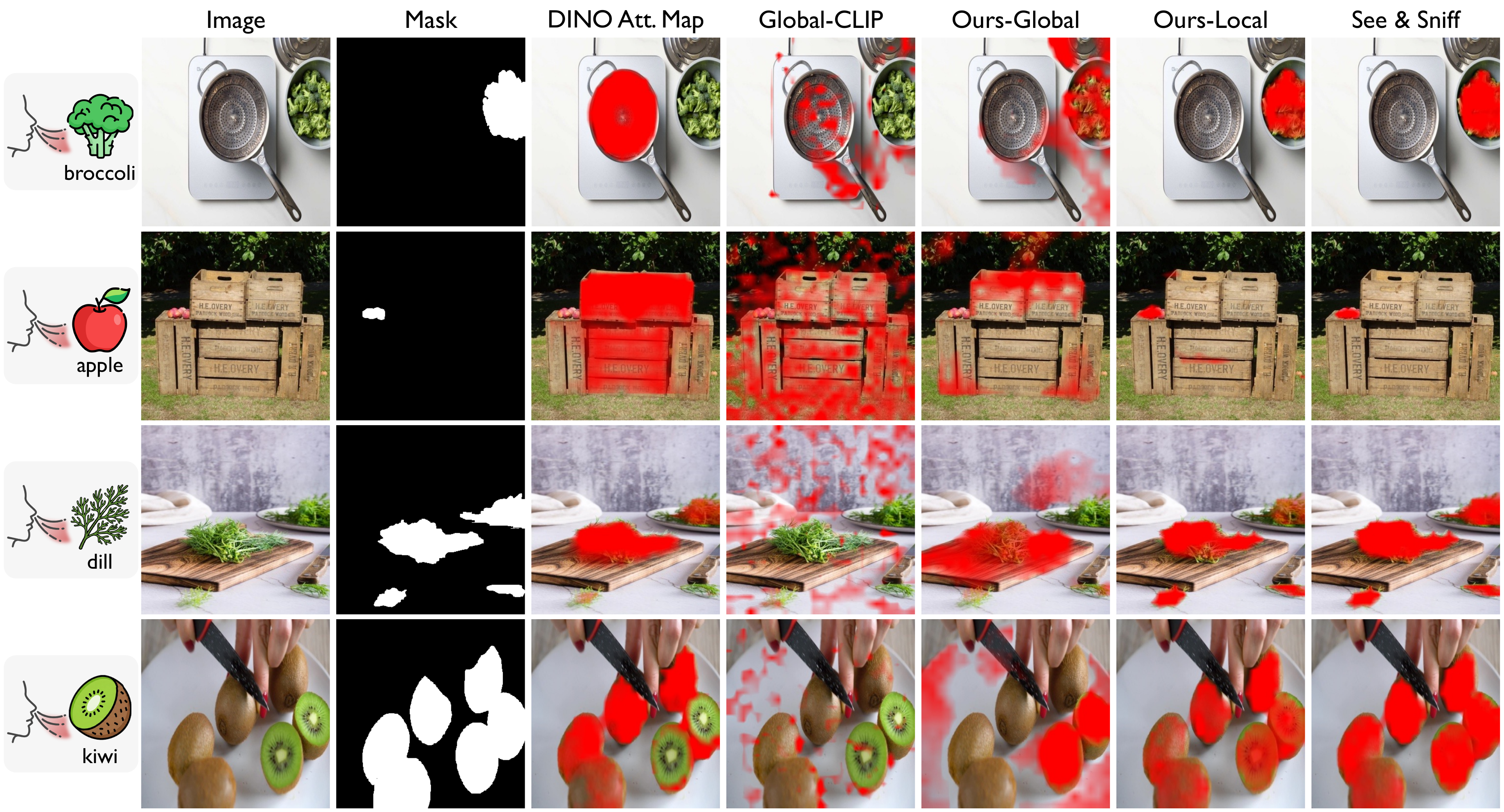}
    \caption{\textbf{Qualitative Smell Localization Results.}}
    \label{fig:localization}
    \vspace{-4mm}
\end{figure*}

\noindent\textbf{-- Smell Localization Task}

\noindent We introduce smell localization as a new task in machine olfaction, aiming to spatially ground a given sniff within a visual scene. Given an olfactory input, the objective is to localize the image region expected to emit the corresponding smell. \textit{SmellNet-V}-Source is used as test set for this task. We use mAP and mIoU as evaluation metrics by following standard multimodal grounding protocols ~\cite{everingham2015pascal,luddecke2022image,cheng2022masked,hamilton2024separating,ryu2025seeing}. Results are in~\Tref{tab:localization} and~\Fref{fig:localization}. \textbf{Key findings} are as follows:

\clearpage

\begin{wraptable}{l}{0.49\columnwidth} 
\vspace{2mm}
\centering
\footnotesize
\setlength{\tabcolsep}{5pt}
\renewcommand{\arraystretch}{1.05}
\vspace{-2mm}
\adjustbox{max width=\linewidth}{
\begin{tabular}{lcc}
\toprule
\textbf{Method} & \textbf{mAP} & \textbf{mIoU} \\
\midrule

\rowcolor{gray!15}
\multicolumn{3}{l}{\textit{Binary mask}} \\
Full square   & --     & 0.2072 \\
Full circle   & --     & 0.2393 \\
\midrule

\rowcolor{gray!15}
\multicolumn{3}{l}{\textit{Visual heatmap}} \\
DINOv3 Att. Map         & 0.6596 & 0.5157 \\
\midrule

\rowcolor{gray!15}
\multicolumn{3}{l}{\textit{Global alignment}} \\
Global-CLIP   & 0.1736 & 0.2073 \\
Ours-Global   & 0.6676 & 0.5016 \\
\midrule

\rowcolor{gray!15}
\multicolumn{3}{l}{\textit{Local alignment}} \\
Ours-Local                 & \underline{0.7970} & \underline{0.6099} \\
\rowcolor{azure!10}
See \& Sniff      & \textbf{0.8362} & \textbf{0.6456} \\
\midrule

\rowcolor{gray!15}
\multicolumn{3}{l}{\textit{Cascaded}} \\
SmellNet + SAM3             & -- & 0.3214 \\
SmellNet + SAM3$^{\dagger}$ & -- & 0.3683 \\
GT + SAM3                    & -- & 0.5854 \\
GT + SAM3$^{\dagger}$        & -- & 0.6700 \\
\bottomrule
\end{tabular}
}
\vspace{-1em}
\caption{\textbf{Smell Localization Results.}}
\label{tab:localization}
\vspace{-2.5em}
\end{wraptable}

\noindent (1) \textit{Local alignment is essential for spatial smell grounding.} Local alignment substantially outperforms global alignment in smell localization. While global embeddings capture holistic correspondence, they fail to model spatial cross-modal interactions. In contrast, our dense local alignment objective learns fine-grained visuo-olfactory correspondences, enabling accurate smell source grounding. This directly validates our architectural design and learning objective.

\noindent (2) \textit{Joint multimodal learning outperforms cascaded pipelines.} The cascaded baseline (SmellNet + SAM3) performs significantly worse than our unified model. This two-stage pipeline lacks direct visuo-olfactory feature interaction and cannot model spatial correspondences during representation learning. These results demonstrate that joint end-to-end multimodal learning is more effective than post-hoc category-conditioned segmentation.

\noindent (3) \textit{Competitive performance against the GT upper-bound baseline.} While GT + SAM3 serves as an upper-bound reference by using ground-truth category labels, SAM3 exhibits inherent limitations for certain smell categories due to out-of-domain effects. To ensure fairness, we additionally report results excluding 7 categories where SAM3 fails to produce meaningful segmentations (marked with $\dagger$). Even under this favorable setting, our model remains competitive and surpasses it without this adjustment, demonstrating the robustness of the learned visuo-olfactory representations. 

\noindent (4) \textit{Visuo-olfactory alignment is necessary for smell localization.} Vision-only baselines such as DINOv3 Att. Maps fail to match our model, indicating that generic visual objectness is insufficient for smell localization. Accurate grounding requires explicit visuo-olfactory alignment rather than purely visual cues.

\noindent (5) \textit{The aligner module further enhances cross-modal alignment.} Incorporating the aligner improves the performance of the local alignment variant (Ours-Local \vs See \& Sniff). This improvement suggests that projecting modality-specific features into a common embedding space strengthens visuo-olfactory correspondence, leading to better smell localization.

\noindent\textbf{Qualitative Results.} The results are shown in ~\Fref{fig:localization}. \textit{See \& Sniff} accurately localizes the target source even when it is off-center, or when multiple source objects are present. For example in the second row, while other baselines focus on the visually salient wooden boxes, our local alignment models correctly identify the apple on the left.

\noindent\textbf{Interactive Localization.}
To further examine fine-grained visuo-olfactory alignment, we evaluate our model under an interactive localization setting following~\cite{senocak2025toward}. In this setup, a single image containing multiple ingredients is paired with different smell inputs, and a reliable localization method should shift its predicted region according to the given smell. This analysis verifies that localization is driven by smell semantics rather than fixed visual saliency.

\begin{figure*}[t!]
    \centering
    \includegraphics[width=0.95\linewidth]{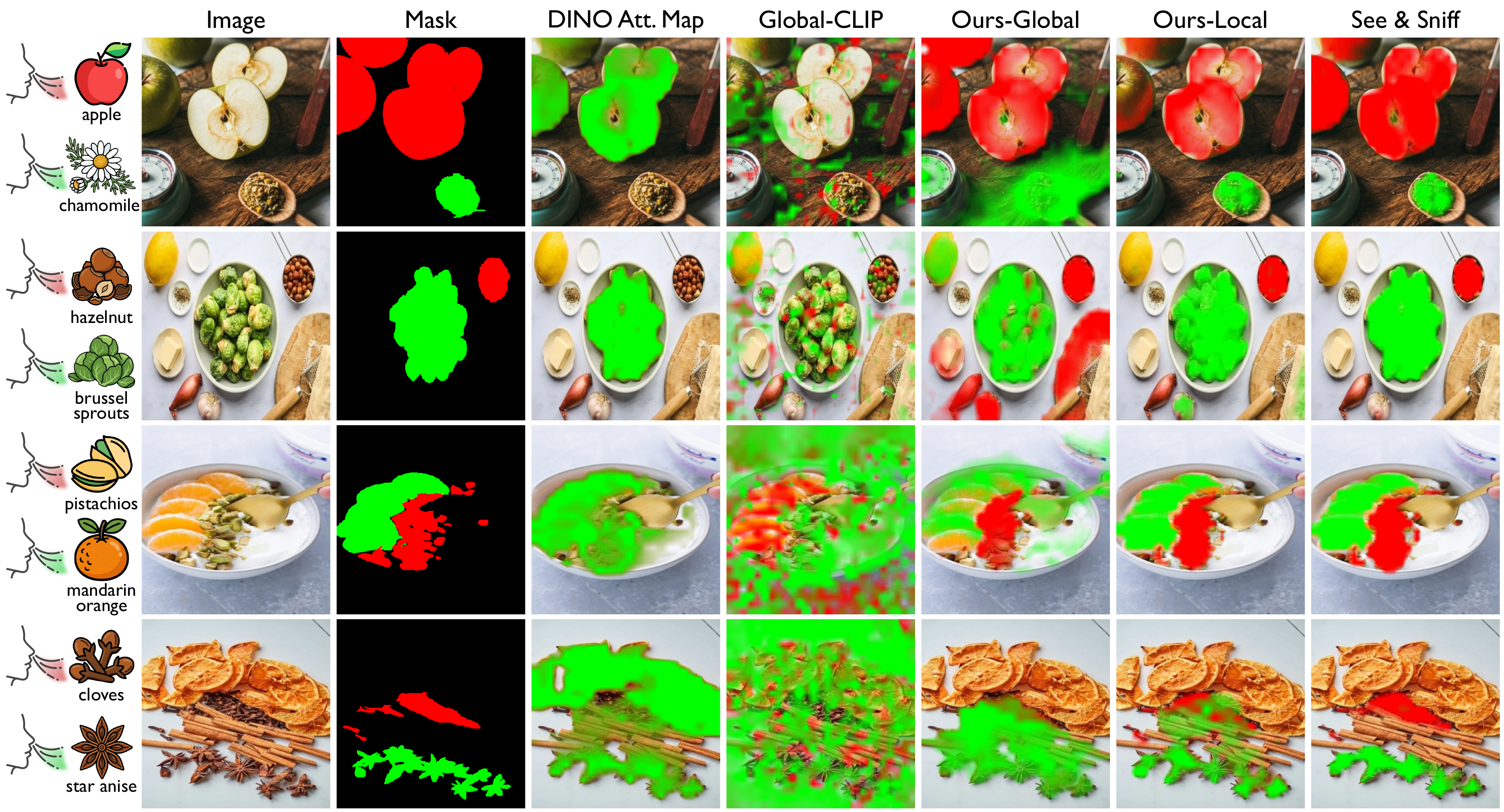}
    \vspace{-2mm}
    \caption{\textbf{Qualitative Results on Interactive Localization.}}
    \label{fig:iiou}
    \vspace{-8mm}
\end{figure*}

\begin{wraptable}{r}{0.42\columnwidth}
\vspace{-2.2em}
\centering
\small
\setlength{\tabcolsep}{7pt}
\begin{tabular}{lc}
\toprule
\textbf{Model} & \textbf{IIoU} \\
\midrule
\rowcolor{gray!15}
\multicolumn{2}{l}{\textit{Visual heatmap}} \\
DINOv3 Att.\ Map & 0.01 \\
\midrule
\rowcolor{gray!15}
\multicolumn{2}{l}{\textit{Global alignment}} \\
Global-CLIP & 0.00 \\
Ours-Global & 0.22 \\
\midrule
\rowcolor{gray!15}
\multicolumn{2}{l}{\textit{Local alignment}} \\
Ours-Local & \underline{0.32} \\
\rowcolor{azure!10}
See \& Sniff & \textbf{0.35} \\
\midrule
\rowcolor{gray!15}
\multicolumn{2}{l}{\textit{Cascaded}} \\
SmellNet + SAM3 & 0.06 \\
GT + SAM3 & 0.41 \\
\bottomrule
\vspace{-4mm}
\end{tabular}

\vspace{-0.9em}
\caption{\textbf{Interactive Localization Results.}}
\label{tab:iiou}
\vspace{-2.5em}
\end{wraptable}

\noindent We construct a new dataset, \textit{SmellNet-V}-InteractiveSource, for interactive smell localization. It contains 100 images, each depicting two distinct ingredients with pixel-level segmentation masks. Forty images are drawn from the \textit{SmellNet-V}-Source test set, while the remaining are newly collected from the web to ensure the presence of multiple ingredients within a single scene. For evaluation, each smell signal is divided into multiple sniff units as before. For a given image, localization performance is computed independently for each sniff unit corresponding to each ingredient, and the IoU scores are averaged to obtain a final score per ingredient. A sample is considered successful only if the IoU for both ingredients exceeds 0.5, ensuring that the model correctly localizes each smell source within the same scene. As shown in~\Tref{tab:iiou}, interactive localization further shows the importance of local cross-modal alignment. 
Global-alignment methods perform poorly, with Global-CLIP failing entirely and Ours-Global achieving only limited success. DINOv3 attention maps also struggle, as they inherently produce a single dominant region. 

\begin{wrapfigure}{r}{0.34\columnwidth} 
\centering
\includegraphics[width=\linewidth]{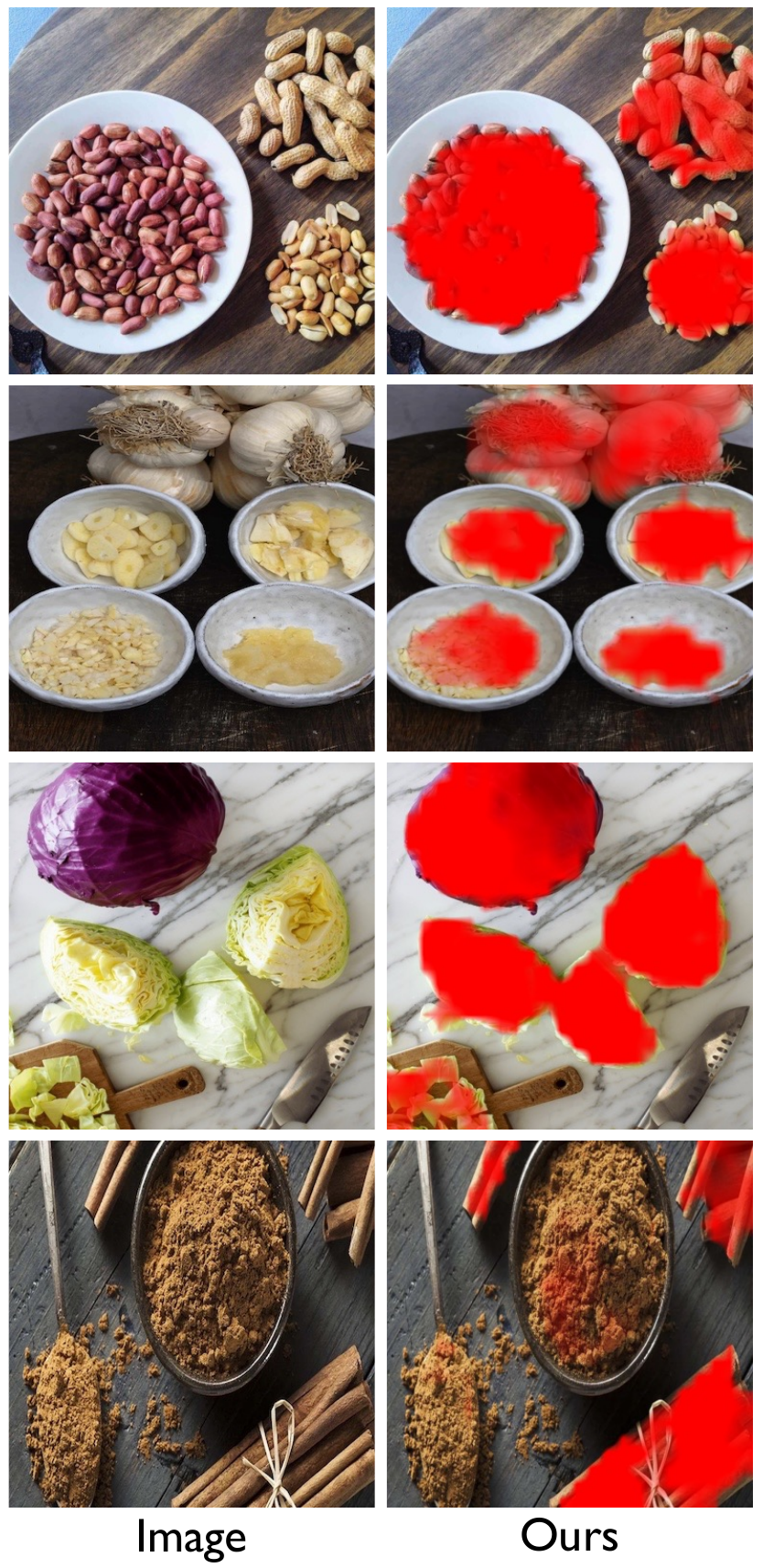}
\vspace{-2em}
\caption{\textbf{Localization across Physical States.}}
\label{fig:notable_cases}
\vspace{-2em}
\end{wrapfigure}

\noindent In contrast, \textit{See \& Sniff}
outperforms all baselines except the upper-bound setting. These findings indicate that interactive smell grounding requires fine-grained, smell-conditioned spatial reasoning rather than static saliency or global embedding similarity. Although the overall IoU values reflect the difficulty of the task, our model consistently shows superior conditional grounding, as further illustrated in~\Fref{fig:iiou}.

\noindent\textbf{Other Discussion.} Lastly, we analyze localization under different physical states of the same ingredient, for example, peanuts with varying shell conditions; peeled, sliced, or crushed garlic; whole or sliced cabbage with varying colors; or ground versus non-ground cinnamon. Qualitative results are shown in~\Fref{fig:notable_cases}. 
Our model successfully localizes peanuts, garlic and cabbage across diverse visual appearances within the same scene, demonstrating robustness to state-level variations.
However, we observe failure cases for powdered spices, as the nearly identical brown, fine-grained textures of different ingredients within the family limit discriminative cues for localization. These findings suggest that while the model is robust to structural variations, texture-dominated states remain challenging due to limited visual separability.

\vspace{-0.6em}
\section{Conclusion and Discussion}
\vspace{-0.6em}
In this work, we take an early step toward visuo-olfactory learning by transforming a unimodal smell dataset into a cross-modal training paradigm through synthetic pairing with in-the-wild images, forming \textit{SmellNet-V}, and by introducing \textit{See \& Sniff}, a self-supervised model with dense local alignment for joint representation learning. Our results show that visual supervision not only enables cross-modal retrieval and smell localization, but also strengthens intrinsic olfactory representations, surpassing smell-only learning by a significant margin in classification. By establishing new benchmarks for classification, retrieval, and spatial grounding, we move machine olfaction beyond unimodal modeling and open a path toward integrating smell into future multimodal perception systems.

\noindent\textbf{What can be further done?} Future work may extend visuo-olfactory learning beyond normal, non-degraded ingredient states to scenarios involving spoilage or transformation, where visual and olfactory cues encode temporal semantic changes (\eg, fermentation or decay). Another promising direction is modeling compositional and mixture odors, enabling recognition and localization of complex food items or multi-ingredient scenes (\eg, lemon cheesecake conditioned on lemon scent). Such extensions would move toward more realistic and semantically rich visuo-olfactory perception.

\section{Acknowledgment}
This work was supported by IITP grants funded by the Korean government (MSIT)
(RS-2024-00457882, National AI Research Lab Project, 50\%;
RS-2020-II201336, Artificial Intelligence Graduate School Program, UNIST, 10\%);
the National Research Foundation of Korea (NRF) grant funded by the Korean government (MSIT)
(RS-2026-25496684, 35\%); and the Hankuk University of Foreign Studies Research Fund (of 2026, 5\%).


%
%
\bibliographystyle{splncs04}
\bibliography{main}

\begin{thebibliography}{10}
\providecommand{\url}[1]{\texttt{#1}}
\providecommand{\urlprefix}{URL }
\providecommand{\doi}[1]{https://doi.org/#1}

\bibitem{abid2019gradio}
Abid, A., Abdalla, A., Abid, A., Khan, D., Alfozan, A., Zou, J.: Gradio: Hassle-free sharing and testing of ml models in the wild. arXiv preprint arXiv:1906.02569  (2019)

\bibitem{achebouche2022application}
Achebouche, R., Tromelin, A., Audouze, K., Taboureau, O.: Application of artificial intelligence to decode the relationships between smell, olfactory receptors and small molecules. Scientific reports  (2022)

\bibitem{arandjelovic2017look}
Arandjelovic, R., Zisserman, A.: Look, listen and learn. In: ICCV (2017)

\bibitem{aytar2016soundnet}
Aytar, Y., Vondrick, C., Torralba, A.: Soundnet: Learning sound representations from unlabeled video. In: NeurIPS (2016)

\bibitem{ba2016layernormalization}
Ba, J.L., Kiros, J.R., Hinton, G.E.: Layer normalization. arXiv preprint arXiv:1607.06450  (2016)

\bibitem{carion2025sam}
Carion, N., Gustafson, L., Hu, Y.T., Debnath, S., Hu, R., Suris, D., Ryali, C., Alwala, K.V., Khedr, H., Huang, A., et~al.: Sam 3: Segment anything with concepts. In: ICLR (2026)

\bibitem{castellotti2025visual}
Castellotti, S., Soldo, M., Plank, T., Viva, M.M.D., Greenlee, M.W.: Visual search performance depends on the congruency of olfactory sensations. Scientific Reports  (2025)

\bibitem{cheng2022masked}
Cheng, B., Misra, I., Schwing, A.G., Kirillov, A., Girdhar, R.: Masked-attention mask transformer for universal image segmentation. In: CVPR (2022)

\bibitem{elizalde2022clap}
Elizalde, B., Deshmukh, S., Ismail, M.A., Wang, H.: Clap: Learning audio concepts from natural language supervision. In: ICASSP (2023)

\bibitem{everingham2015pascal}
Everingham, M., Eslami, S.A., Van~Gool, L., Williams, C.K., Winn, J., Zisserman, A.: The pascal visual object classes challenge: A retrospective. IJCV  (2015)

\bibitem{fancyfeast2025watermark}
{Fancy Feast}: {Joycaption Watermark Detection}. {Hugging Face Spaces} (2025), \url{https://huggingface.co/spaces/fancyfeast/joycaption-watermark-detection}, {Accessed} 24 June 2026

\bibitem{feng2025smellnet}
Feng, D., Dai, W., Li, C., Pernigo, A., Wen, Y., Liang, P.P.: Smellnet: A large-scale dataset for real-world smell recognition. In: ICLR (2026)

\bibitem{fichtelmann2025machine}
Fichtelmann, P., Westermayr, J.: Machine learning for smell: Ordinal odor strength prediction of molecular perfumery components. arXiv preprint arXiv:2512.08683  (2025)

\bibitem{fu2024a}
Fu, L., Datta, G., Huang, H., Panitch, W.C.H., Drake, J., Ortiz, J., Mukadam, M., Lambeta, M., Calandra, R., Goldberg, K.: A touch, vision, and language dataset for multimodal alignment. In: ICML (2024)

\bibitem{girdhar2023imagebind}
Girdhar, R., El-Nouby, A., Liu, Z., Singh, M., Alwala, K.V., Joulin, A., Misra, I.: Imagebind: One embedding space to bind them all. In: CVPR (2023)

\bibitem{gong2022contrastive}
Gong, Y., Rouditchenko, A., Liu, A.H., Harwath, D., Karlinsky, L., Kuehne, H., Glass, J.: Contrastive audio-visual masked autoencoder. In: ICLR (2022)

\bibitem{hamilton2024separating}
Hamilton, M., Zisserman, A., Hershey, J.R., Freeman, W.T.: Separating the" chirp" from the" chat": Self-supervised visual grounding of sound and language. In: CVPR (2024)

\bibitem{harwath2018jointly}
Harwath, D., Recasens, A., Sur{\'\i}s, D., Chuang, G., Torralba, A., Glass, J.: Jointly discovering visual objects and spoken words from raw sensory input. In: ECCV (2018)

\bibitem{cvat}
{Intel Corporation}: {CVAT: Computer Vision Annotation Tool} (2025), \url{https://www.cvat.ai/}, {Accessed} 24 June 2026

\bibitem{iwata2025interpretable}
Iwata, H.: Interpretable multitask deep learning models for odor perception based on molecular structure. Current Research in Food Science  (2025)

\bibitem{jia2021scaling}
Jia, C., Yang, Y., Xia, Y., Chen, Y.T., Parekh, Z., Pham, H., Le, Q., Sung, Y.H., Li, Z., Duerig, T.: Scaling up visual and vision-language representation learning with noisy text supervision. In: ICML (2021)

\bibitem{kang2026aromma}
Kang, D., Kim, J., Park, J., Lee, K., Choi, J.W., So, J.: Aromma: Unifying olfactory embeddings for single molecules and mixtures. arXiv preprint arXiv:2601.19561  (2026)

\bibitem{keller1934neglected}
Keller, H.: A neglected treasure. The Home Magazine  (1934), \url{https://www.afb.org/HelenKellerArchive?a=d&d=A-HK02-B225-F02-024}

\bibitem{kim2026seeing}
Kim, S., Lee, S., Ryu, H., Chung, J.S., Senocak, A.: Seeing through touch: Tactile-driven visual localization of material regions. In: CVPR (2026)

\bibitem{kirillov2023segment}
Kirillov, A., Mintun, E., Ravi, N., Mao, H., Rolland, C., Gustafson, L., Xiao, T., Whitehead, S., Berg, A.C., Lo, W.Y., et~al.: Segment anything. In: ICCV (2023)

\bibitem{lee2023principal}
Lee, B.K., Mayhew, E.J., Sanchez-Lengeling, B., Wei, J.N., Qian, W.W., Little, K.A., Andres, M., Nguyen, B.B., Moloy, T., Yasonik, J., et~al.: A principal odor map unifies diverse tasks in olfactory perception. Science  (2023)

\bibitem{li2023blip}
Li, J., Li, D., Savarese, S., Hoi, S.: Blip-2: Bootstrapping language-image pre-training with frozen image encoders and large language models. In: ICML (2023)

\bibitem{li2022blip}
Li, J., Li, D., Xiong, C., Hoi, S.: Blip: Bootstrapping language-image pre-training for unified vision-language understanding and generation. In: ICML (2022)

\bibitem{luddecke2022image}
L{\"u}ddecke, T., Ecker, A.: Image segmentation using text and image prompts. In: CVPR (2022)

\bibitem{lyu2024omnibind}
Lyu, Y., Zheng, X., Kim, D., Wang, L.: Omnibind: Teach to build unequal-scale modality interaction for omni-bind of all. arXiv preprint arXiv:2405.16108  (2024)

\bibitem{mainland2006sniff}
Mainland, J., Sobel, N.: The sniff is part of the olfactory percept. Chemical senses  (2006)

\bibitem{mueller2019scent}
Mueller, P., Salminen, K., Nieminen, V., Kontunen, A., Karjalainen, M., Isokoski, P., Rantala, J., Savia, M., V{\"a}liaho, J., Kallio, P., et~al.: Scent classification by k nearest neighbors using ion-mobility spectrometry measurements. Expert systems with applications  (2019)

\bibitem{naeem2024silc}
Naeem, M.F., Xian, Y., Zhai, X., Hoyer, L., Van~Gool, L., Tombari, F.: Silc: Improving vision language pretraining with self-distillation. In: ECCV (2024)

\bibitem{owens2018audio}
Owens, A., Efros, A.A.: Audio-visual scene analysis with self-supervised multisensory features. In: ECCV (2018)

\bibitem{ozguroglu2025new}
Ozguroglu, E., Liang, J., Liu, R., Chiquier, M., DeTienne, M., Qian, W.W., Horowitz, A., Owens, A., Vondrick, C.: New york smells: A large multimodal dataset for olfaction. arXiv preprint arXiv:2511.20544  (2025)

\bibitem{radford2021learning}
Radford, A., Kim, J.W., Hallacy, C., Ramesh, A., Goh, G., Agarwal, S., Sastry, G., Askell, A., Mishkin, P., Clark, J., et~al.: Learning transferable visual models from natural language supervision. In: ICML (2021)

\bibitem{raithel2021using}
Raithel, C.U., Gottfried, J.A.: Using your nose to find your way: Ethological comparisons between human and non-human species. Neuroscience \& Biobehavioral Reviews  (2021)

\bibitem{ryu2025seeing}
Ryu, H., Kim, S., Chung, J.S., Senocak, A.: Seeing speech and sound: Distinguishing and locating audio sources in visual scenes. In: CVPR (2025)

\bibitem{sanchez2019machine}
Sanchez-Lengeling, B., Wei, J.N., Lee, B.K., Gerkin, R.C., Aspuru-Guzik, A., Wiltschko, A.B.: Machine learning for scent: Learning generalizable perceptual representations of small molecules. arXiv preprint arXiv:1910.10685  (2019)

\bibitem{senocak2018learning}
Senocak, A., Oh, T.H., Kim, J., Yang, M.H., Kweon, I.S.: Learning to localize sound source in visual scenes. In: CVPR (2018)

\bibitem{senocak2023sound}
Senocak, A., Ryu, H., Kim, J., Oh, T.H., Pfister, H., Chung, J.S.: Sound source localization is all about cross-modal alignment. In: ICCV (2023)

\bibitem{senocak2025toward}
Senocak, A., Ryu, H., Kim, J., Oh, T.H., Pfister, H., Chung, J.S.: Toward interactive sound source localization: Better align sight and sound! IEEE TPAMI  (2025)

\bibitem{simeoni2025dinov3}
Sim{\'e}oni, O., Vo, H.V., Seitzer, M., Baldassarre, F., Oquab, M., Jose, C., Khalidov, V., Szafraniec, M., Yi, S., Ramamonjisoa, M., et~al.: Dinov3. arXiv preprint arXiv:2508.10104  (2025)

\bibitem{singh2025openai}
Singh, A., Fry, A., Perelman, A., Tart, A., Ganesh, A., El-Kishky, A., McLaughlin, A., Low, A., Ostrow, A., Ananthram, A., et~al.: Openai gpt-5 system card. arXiv preprint arXiv:2601.03267  (2025)

\bibitem{sung2024data}
Sung, S.H., Suh, J.M., Hwang, Y.J., Jang, H.W., Park, J.G., Jun, S.C.: Data-centric artificial olfactory system based on the eigengraph. Nature communications  (2024)

\bibitem{tran2019deepnose}
Tran, N., Kepple, D., Shuvaev, S., Koulakov, A.: Deepnose: Using artificial neural networks to represent the space of odorants. In: ICML (2019)

\bibitem{vaswani2023attentionneed}
Vaswani, A., Shazeer, N., Parmar, N., Uszkoreit, J., Jones, L., Gomez, A.N., Kaiser, L., Polosukhin, I.: Attention is all you need. In: NeurIPS (2017)

\bibitem{wachowiak2011all}
Wachowiak, M.: All in a sniff: olfaction as a model for active sensing. Neuron  (2011)

\bibitem{wu2018unsupervised}
Wu, Z., Xiong, Y., Yu, S., Lin, D.: Unsupervised feature learning via non-parametric instance-level discrimination. In: CVPR (2018)

\bibitem{yang2024binding}
Yang, F., Feng, C., Chen, Z., Park, H., Wang, D., Dou, Y., Zeng, Z., Chen, X., Gangopadhyay, R., Owens, A., et~al.: Binding touch to everything: Learning unified multimodal tactile representations. In: CVPR (2024)

\bibitem{yang2022touch}
Yang, F., Ma, C., Zhang, J., Zhu, J., Yuan, W., Owens, A.: Touch and go: Learning from human-collected vision and touch. In: NeurIPS - Datasets and Benchmarks Track (2022)

\bibitem{yang2023generating}
Yang, F., Zhang, J., Owens, A.: Generating visual scenes from touch. In: ICCV (2023)

\bibitem{zhai2023sigmoid}
Zhai, X., Mustafa, B., Kolesnikov, A., Beyer, L.: Sigmoid loss for language image pre-training. In: ICCV (2023)

\end{thebibliography}

\definecolor{positive}{RGB}{44,107,190}
\definecolor{negative}{RGB}{204,41,41}

\clearpage

\begin{center}
{\Large\bfseries -- Supplementary Material --}\\[0.35em]
{\Large\bfseries See \& Sniff: Learning Visuo-Olfactory\\ Representations}
\end{center}

\vspace{2.5em}

\setcounter{section}{6}

\startcontents[supp]

\vspace{1em}
\noindent The contents in this supplementary material are as follows:
\providecommand{\authcount}[1]{}
\providecommand{\orcidID}[1]{}
\vspace{0.5em}

\begingroup
    \hypersetup{linkcolor=black}
    \renewcommand{\contentsname}{}
    \renewcommand{\clearpage}{}
    \renewcommand{\newpage}{}
    \printcontents[supp]{}{1}[1]{}
\endgroup

\vspace{0.3em}
\noindent\rule{\linewidth}{0.2pt}

\section{Details on \textit{SmellNet-V}}\label{sec:datasets}

To construct \textit{SmellNet-V}, we curate semantically aligned in-the-wild web images through a systematic pipeline of image collection and filtering. This creates multimodal data that closely approximates natural visuo-olfactory correspondences.

\noindent\textbf{Image Collection.} To ensure the visual robustness of \textit{SmellNet-V}, we utilize an LLM~\cite{singh2025openai} as an expert curator to synthesize diverse search queries for each ingredient. SmellNet provides textual olfactory descriptions for each ingredient. Our curator incorporates the original SmellNet ingredient name together with descriptive text to better match each target olfactory profile. This contextual grounding prevents semantic ambiguity when an ingredient possesses multiple visual forms. For instance, we specifically select coriander ``seeds’’ (rather than ``leaves’’) as defined in the description, or identify angelica ``roots’’ to match ``earthy'' and ``woody'' scent descriptors.

\begin{figure*}[t!]
    \centering
    \includegraphics[width=0.9\linewidth]{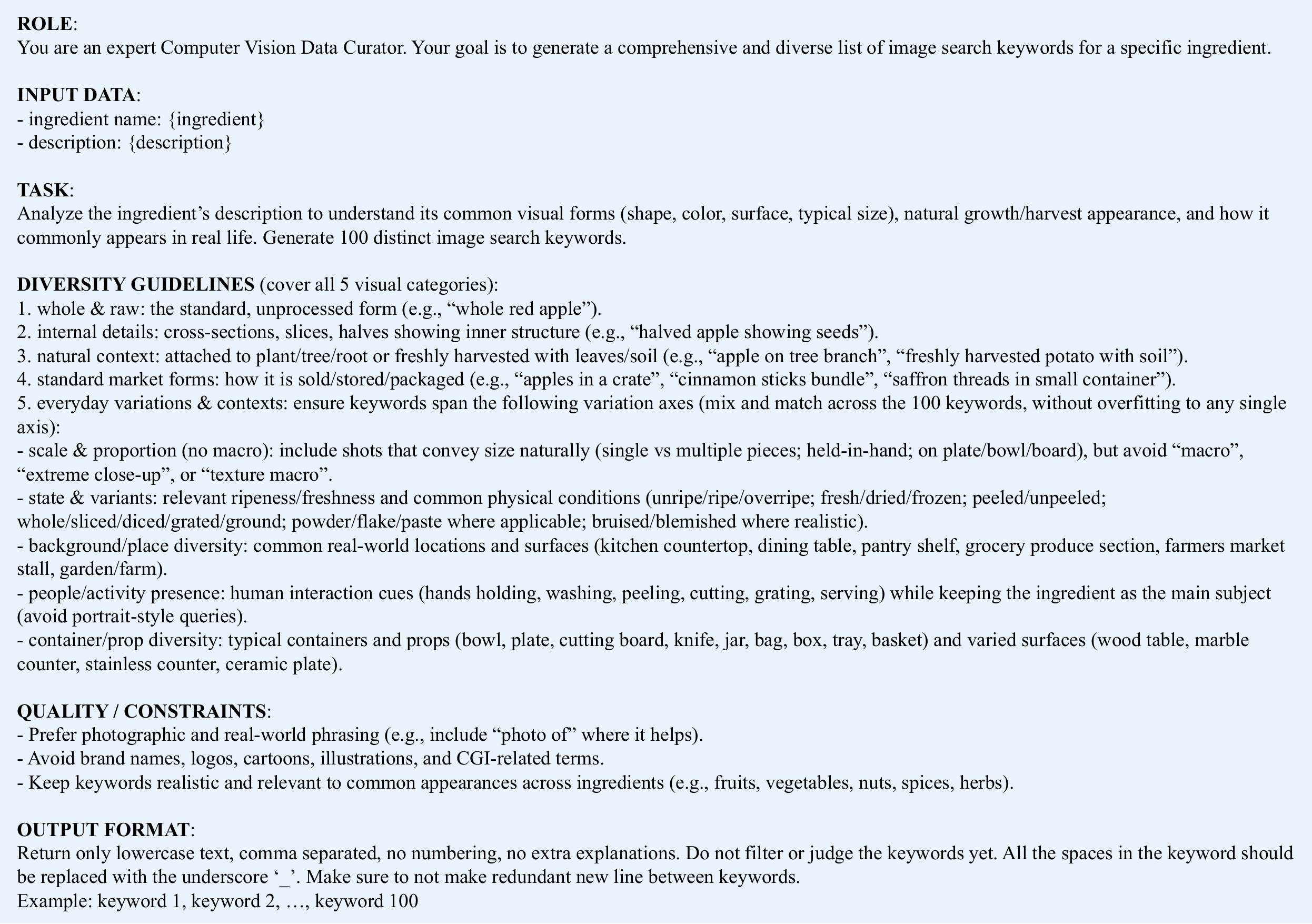}
    \vspace{-0.8em}
    \caption{\textbf{Prompt for Search Query Generation.}}
    \label{fig:suppl_query_prompt}
    \vspace{-7mm}
\end{figure*}

As detailed in~\Fref{fig:suppl_query_prompt}, these queries are systematically generated across five key visual dimensions: (i) whole and raw forms, (ii) internal details (\eg, cross-sections and slices), (iii) natural growth (\eg, attached to branches or freshly harvested), (iv) market forms (\eg, in crates or jars), and (v) everyday variations (spanning diverse camera scales, physical conditions, environmental backgrounds, human interactions, and container types). For instance, the ingredient ``apple'' returns queries like `\textit{whole apple on wooden table}', `\textit{apples in grocery produce section}', and `\textit{apple hanging on tree}'.

\begin{figure*}[h!]
    \centering
    \includegraphics[width=0.98\linewidth]{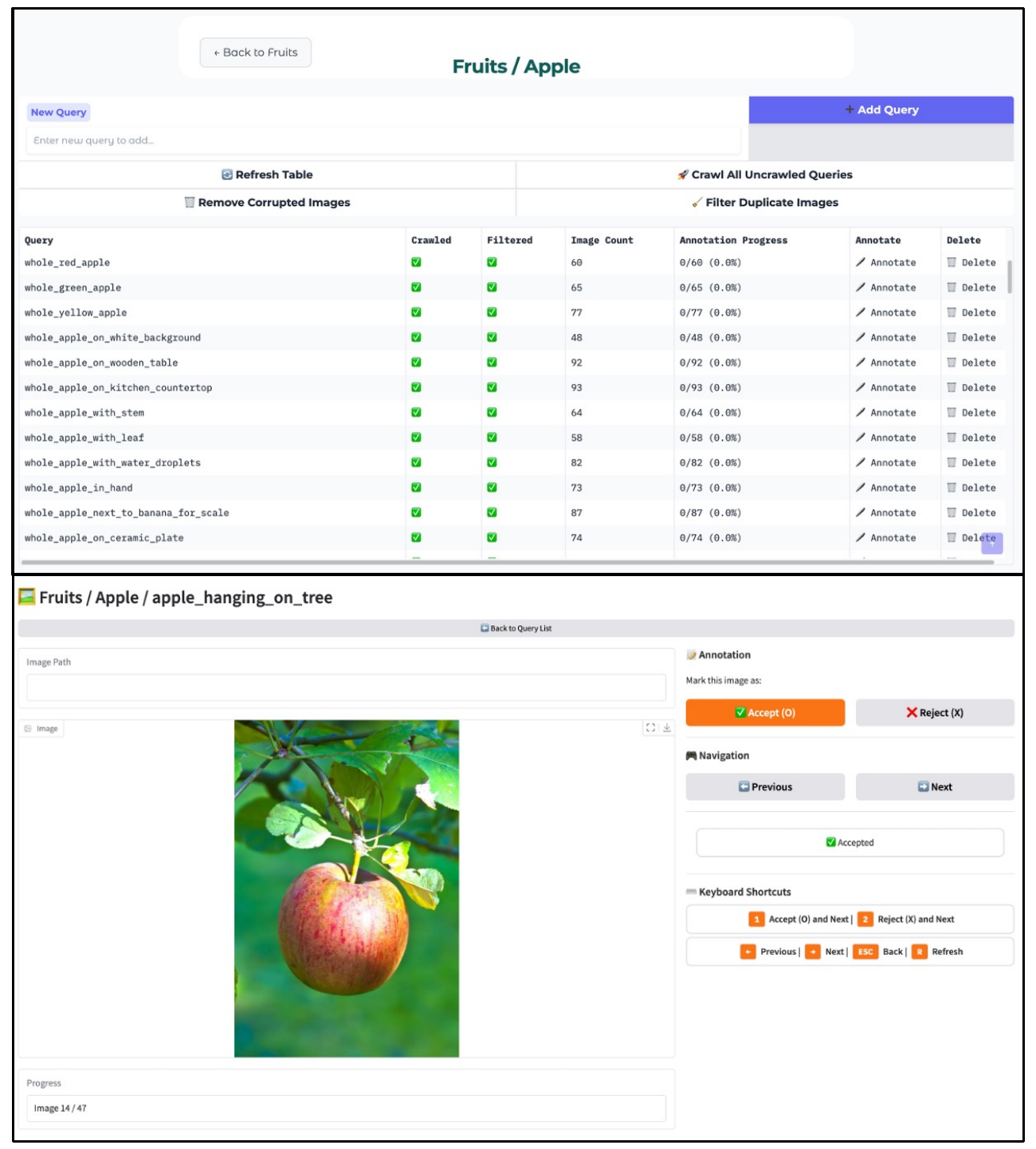}
    \vspace{-0.5em}
    \caption{\textbf{Custom-built Gradio Pages for Image Collection (Top) and Image Filtering (Bottom).}}
    \label{fig:suppl_gradio}
    \vspace{-4mm}
\end{figure*}

To manage this high-volume data pipeline, we developed a custom Gradio-based~\cite{abid2019gradio} curation tool (\Fref{fig:suppl_gradio}, Top), which facilitates ingredient-specific query refinement, automated image crawling, and the systematic removal of corrupted or duplicate samples to ensure the integrity of the final collection.

\noindent\textbf{Image Filtering.} To ensure semantic precision and visual high-fidelity, we implement a rigorous three-stage filtering pipeline: (1) \textit{prompt-based verification}, (2) \textit{quality filtering}, and (3) \textit{human refinement}.

\begin{wrapfigure}{r}{0.34\columnwidth} 
\vspace{-2.3em}
\centering
\includegraphics[width=\linewidth]{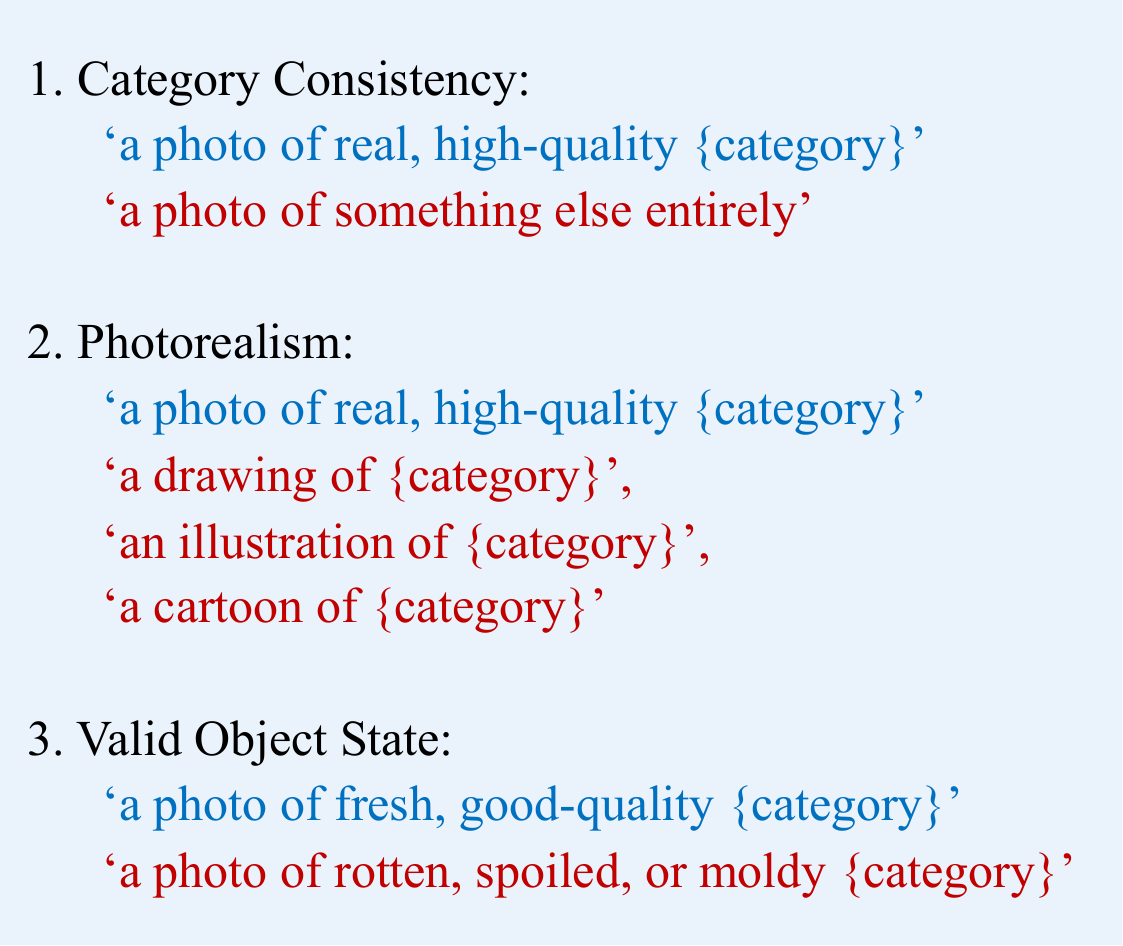}
\vspace{-1.6em}
\caption{\textbf{\textcolor{positive}{Positive} and \textcolor{negative}{Negative} prompt pairs for CLIP verification.}}
\label{fig:suppl_clip_prompt}
\vspace{-2em}
\end{wrapfigure}

First, we perform \textit{prompt-based verification} using CLIP~\cite{radford2021learning} to validate candidates across three criteria: category consistency, photorealism, and object state. In particular, the state-based filter excludes degraded or spoiled instances, ensuring the data aligns with the standard, non-degraded states of ingredients typically sourced from public retailers as noted in SmellNet~\cite{feng2025smellnet}. As illustrated in~\Fref{fig:suppl_clip_prompt}, we employ contrastive prompt pairs for each criterion, retaining an image only if its similarity score for the positive prompt exceeds those of all corresponding negative prompts.

Next, we apply \textit{quality filtering} to eliminate images with excessive watermarks. Samples containing five or more detected watermarks (by using an off-the-shelf watermark detector~\cite{fancyfeast2025watermark}) are automatically discarded to preserve visual integrity. Finally, \textit{human refinement} is conducted via a custom Gradio-based tool (\Fref{fig:suppl_gradio}, Bottom). Annotators manually review the remaining images to prune any irrelevant or low-quality samples, finalizing a highly curated collection.

\newpage
\section{Implementation Details}\label{sec:implement}
\noindent\textbf{Training Details.} We train the model for 500 epochs with a batch size of 64, using the AdamW optimizer $(\beta_{1}=0.9,\beta_{2}=0.999)$ with a learning rate of $5\times10^{-5}$ and a weight decay of 0.05. The dimensionality of each embedding from the modality backbones and aligners is set to 384. Following~\cite{feng2025smellnet}, we utilize six stable sensors ($\mathrm{NO_2, C_2H_5OH, VOC, CO, Alcohol,LPG}$) out of 12 available sensors, excluding Benzene, Temperature, Pressure, Humidity, Gas Resistance, and Altitude. In our default setting, the first-order temporal difference (lag), window size, and stride are set to $p=25$, $W=50$, and $s=25$, respectively.

\noindent\textbf{Details of Smell Localization Task.} We summarize the data distribution of the two \textit{SmellNet-V} localization test subsets, \textit{Source} and \textit{InteractiveSource}. 
For the vision modality, \textit{SmellNet-V}-Source contains 20 images for all 50 ingredients, totaling 1,000 images. In \textit{SmellNet-V}-InteractiveSource, each ingredient appears mostly four times. Since each image includes two ingredient masks, this yields 200 ingredient occurrences in total, corresponding to 100 unique images. For the olfactory modality, we use the official SmellNet test split, containing 1,083 sniff units under our default setting. Leveraging our synthetic pairing approach, we evaluate every possible combination between these sniff units and the images of each corresponding ingredient, ensuring a comprehensive assessment across all available pairs for the localization task.

To evaluate the cascaded and upper-bound baselines using SAM3~\cite{carion2025sam} (in Section 4.3 - Smell Localization Task of the main paper), we excluded 7 categories where the SAM3 failed significantly, achieving an IoU of less than 0.1. These categories consist of `allspice', `angelica', `chamomile', `chervil', `cumin', `mugwort' and `pili nut'. The corresponding results in the ~\Tref{tab:localization} of the main paper are marked with $\dagger$ to indicate this exclusion. Note that this exclusion favors these baselines, as our model is evaluated on the full set of ingredients without such filtering.

\section{Ingredient-Wise Results}\label{sec:ingredient_results}

\noindent To complement the results presented in Section 4.3 of the main paper, we report the ingredient-wise performance for smell-classification, cross-modal retrieval, and smell-localization tasks. The detailed results for each task are summarized in~\Tref{tab:suppl_ingredient_classification},~\Tref{tab:suppl_ingredient_retrieval}, and~\Tref{tab:suppl_ingredient_localization} respectively.

In~\Tref{tab:suppl_ingredient_classification}, we compare ingredient-wise smell classification accuracy between \textit{See \& Sniff} and SmellNet.
\textit{See \& Sniff} achieves improvements on 31 out of 50 ingredients. Notably, among 16 challenging ingredients whose SmellNet accuracy is below 0.3, \textit{See \& Sniff} improves performance on 10 ingredients. This highlights the practical impact of our cross-modal alignment, substantially boosting the accuracy for ingredients such as apple and turnip that are nearly indistinguishable under smell-only learning, \ie, unlocking new ingredients.

\begin{table*}[!t]
\centering
\small
\setlength{\tabcolsep}{4pt}
\resizebox{0.65\columnwidth}{!}{
\begin{tabular}{l|l|r|r|r}
\toprule
\textbf{Family} & \textbf{Ingredient} & \textbf{SmellNet} & \textbf{See \& Sniff} & \textbf{$\Delta$} \\
\midrule

\multirow{10}{*}{Fruits}
& Apple            & 4.55  & 77.27 & +72.73 \\
& Banana           & 45.00 & 50.00 & +5.00 \\
& Kiwi             & 45.00 & 55.00 & +10.00 \\
& Lemon            & 71.43 & 47.62 & -23.81 \\
& Mandarin\_Orange & 63.64 & 40.91 & -22.73 \\
& Mango            & 22.73 & 36.36 & +13.64 \\
& Peach            & 68.42 & 78.95 & +10.53 \\
& Pear             & 72.73 & 77.27 & +4.55 \\
& Pineapple        & 90.48 & 95.24 & +4.76 \\
& Strawberry       & 68.00 & 84.00 & +16.00 \\
\midrule

\multirow{10}{*}{Herbs}
& Angelica   & 87.50 & 62.50 & -25.00 \\
& Chamomile  & 22.73 & 22.73 & 0.00 \\
& Chives     & 39.13 & 43.48 & +4.35 \\
& Coriander  & 81.82 & 90.91 & +9.09 \\
& Dill       & 45.00 & 50.00 & +5.00 \\
& Garlic     & 68.18 & 40.91 & -27.27 \\
& Mint       & 5.00  & 0.00  & -5.00 \\
& Mugwort    & 65.00 & 85.00 & +20.00 \\
& Oregano    & 70.00 & 80.00 & +10.00 \\
& Turnip     & 4.55  & 40.91 & +36.36 \\
\midrule

\multirow{10}{*}{Nuts}
& Almond      & 78.26 & 100.00 & +21.74 \\
& Brazil\_Nut & 63.64 & 54.55  & -9.09 \\
& Cashew      & 23.81 & 57.14  & +33.33 \\
& Chestnuts   & 95.45 & 100.00 & +4.55 \\
& Hazelnut    & 23.81 & 23.81  & 0.00 \\
& Peanuts     & 19.05 & 19.05  & 0.00 \\
& Pecans      & 54.55 & 45.45  & -9.09 \\
& Pili\_Nut   & 45.45 & 95.45  & +50.00 \\
& Pistachios  & 83.33 & 95.83  & +12.50 \\
& Walnuts     & 18.75 & 9.38   & -9.38 \\
\midrule

\multirow{10}{*}{Spices}
& Allspice    & 66.67 & 80.95 & +14.29 \\
& Chervil     & 31.82 & 22.73 & -9.09 \\
& Cinnamon    & 28.57 & 52.38 & +23.81 \\
& Cloves      & 90.91 & 90.91 & 0.00 \\
& Cumin       & 81.82 & 90.91 & +9.09 \\
& Ginger      & 23.81 & 47.62 & +23.81 \\
& Mustard     & 13.64 & 13.64 & 0.00 \\
& Nutmeg      & 65.00 & 65.00 & 0.00 \\
& Saffron     & 68.42 & 78.95 & +10.53 \\
& Star\_Anise & 77.27 & 63.64 & -13.64 \\
\midrule

\multirow{10}{*}{Vegetables}
& Asparagus        & 71.43 & 47.62  & -23.81 \\
& Avocado          & 61.90 & 28.57  & -33.33 \\
& Broccoli         & 42.11 & 73.68  & +31.58 \\
& Brussel\_Sprouts & 0.00  & 4.76   & +4.76 \\
& Cabbage          & 20.83 & 37.50  & +16.67 \\
& Cauliflower      & 52.38 & 61.90  & +9.52 \\
& Potato           & 90.00 & 100.00 & +10.00 \\
& Radish           & 95.65 & 95.65  & 0.00 \\
& Sweet\_Potato    & 28.57 & 47.62  & +19.05 \\
& Tomato           & 19.05 & 38.10  & +19.05 \\
\bottomrule
\end{tabular}
}
\vspace{0.5em}
\caption{Ingredient-wise classification accuracy (\%) comparison between SmellNet and \textit{See \& Sniff}. $\Delta$ denotes the performance gain of \textit{See \& Sniff} over SmellNet.}
\label{tab:suppl_ingredient_classification}
\end{table*}
\clearpage
\begin{table*}[!t]
\centering
\small
\setlength{\tabcolsep}{4pt}
\resizebox{0.8\columnwidth}{!}{
\begin{tabular}{l|l|ccc|ccc}
\toprule
\textbf{Family} & \textbf{Ingredient} & \multicolumn{3}{c|}{\textbf{Smell $\rightarrow$ Vision}} & \multicolumn{3}{c}{\textbf{Vision $\rightarrow$ Smell}} \\
 &  & \textbf{R@1} & \textbf{R@5} & \textbf{R@10} & \textbf{R@1} & \textbf{R@5} & \textbf{R@10} \\
\midrule

\multirow{10}{*}{Fruits}
& Apple            & 81.82 & 81.82 & 86.36  & 85.00 & 90.00 & 95.00 \\
& Banana           & 40.00 & 55.00 & 60.00  & 80.00 & 85.00 & 85.00 \\
& Kiwi             & 50.00 & 50.00 & 50.00  & 10.00 & 95.00 & 100.00 \\
& Lemon            & 38.10 & 47.62 & 47.62  & 90.00 & 100.00 & 100.00 \\
& Mandarin\_Orange & 31.82 & 40.91 & 45.45  & 80.00 & 100.00 & 100.00 \\
& Mango            & 36.36 & 36.36 & 36.36  & 15.00 & 60.00 & 100.00 \\
& Peach            & 78.95 & 84.21 & 84.21  & 65.00 & 90.00 & 95.00 \\
& Pear             & 68.18 & 72.73 & 72.73  & 90.00 & 100.00 & 100.00 \\
& Pineapple        & 95.24 & 95.24 & 100.00 & 95.00 & 95.00 & 95.00 \\
& Strawberry       & 84.00 & 84.00 & 88.00  & 40.00 & 100.00 & 100.00 \\
\midrule

\multirow{10}{*}{Herbs}
& Angelica   & 54.17 & 54.17 & 58.33  & 20.00 & 90.00 & 95.00 \\
& Chamomile  & 22.73 & 22.73 & 22.73  & 65.00 & 90.00 & 95.00 \\
& Chives     & 47.83 & 56.52 & 60.87  & 85.00 & 100.00 & 100.00 \\
& Coriander  & 81.82 & 90.91 & 90.91  & 75.00 & 95.00 & 100.00 \\
& Dill       & 55.00 & 60.00 & 65.00  & 60.00 & 90.00 & 90.00 \\
& Garlic     & 40.91 & 40.91 & 40.91  & 85.00 & 90.00 & 90.00 \\
& Mint       & 10.00 & 20.00 & 40.00  & 5.00  & 5.00  & 10.00 \\
& Mugwort    & 85.00 & 90.00 & 90.00  & 90.00 & 95.00 & 100.00 \\
& Oregano    & 70.00 & 80.00 & 80.00  & 80.00 & 85.00 & 90.00 \\
& Turnip     & 40.91 & 45.45 & 50.00  & 35.00 & 65.00 & 70.00 \\
\midrule

\multirow{10}{*}{Nuts}
& Almond      & 100.00 & 100.00 & 100.00 & 45.00 & 70.00 & 75.00 \\
& Brazil\_Nut & 63.64  & 63.64  & 72.73  & 40.00 & 80.00 & 90.00 \\
& Cashew      & 57.14  & 61.90  & 76.19  & 15.00 & 70.00 & 85.00 \\
& Chestnuts   & 95.45  & 100.00 & 100.00 & 70.00 & 90.00 & 95.00 \\
& Hazelnut    & 23.81  & 28.57  & 28.57  & 25.00 & 70.00 & 75.00 \\
& Peanuts     & 14.29  & 19.05  & 28.57  & 30.00 & 80.00 & 85.00 \\
& Pecans      & 36.36  & 45.45  & 63.64  & 65.00 & 100.00 & 100.00 \\
& Pili\_Nut   & 95.45  & 100.00 & 100.00 & 35.00 & 85.00 & 85.00 \\
& Pistachios  & 95.83  & 95.83  & 95.83  & 85.00 & 85.00 & 90.00 \\
& Walnuts     & 9.38   & 12.50  & 15.62  & 65.00 & 80.00 & 90.00 \\
\midrule

\multirow{10}{*}{Spices}
& Allspice   & 85.71 & 85.71 & 85.71 & 85.00  & 100.00 & 100.00 \\
& Chervil    & 13.64 & 36.36 & 36.36 & 85.00  & 95.00  & 95.00 \\
& Cinnamon   & 57.14 & 61.90 & 66.67 & 65.00  & 95.00  & 95.00 \\
& Cloves     & 95.45 & 95.45 & 95.45 & 95.00  & 95.00  & 95.00 \\
& Cumin      & 90.91 & 95.45 & 95.45 & 75.00  & 95.00  & 95.00 \\
& Ginger     & 47.62 & 52.38 & 52.38 & 100.00 & 100.00 & 100.00 \\
& Mustard    & 9.09  & 13.64 & 13.64 & 75.00  & 85.00  & 85.00 \\
& Nutmeg     & 65.00 & 65.00 & 65.00 & 60.00  & 70.00  & 80.00 \\
& Saffron    & 78.95 & 78.95 & 84.21 & 95.00  & 100.00 & 100.00 \\
& Star\_Anise & 59.09 & 63.64 & 63.64 & 90.00  & 90.00  & 100.00 \\
\midrule

\multirow{10}{*}{Vegetables}
& Asparagus        & 61.90 & 66.67 & 71.43  & 90.00 & 100.00 & 100.00 \\
& Avocado          & 23.81 & 33.33 & 42.86  & 15.00 & 90.00  & 100.00 \\
& Broccoli         & 68.42 & 78.95 & 84.21  & 100.00 & 100.00 & 100.00 \\
& Brussel\_Sprouts & 4.76  & 4.76  & 4.76   & 5.00  & 40.00  & 80.00 \\
& Cabbage          & 29.17 & 50.00 & 50.00  & 40.00 & 95.00  & 95.00 \\
& Cauliflower      & 61.90 & 66.67 & 66.67  & 65.00 & 95.00  & 95.00 \\
& Potato           & 100.00 & 100.00 & 100.00 & 95.00 & 95.00  & 100.00 \\
& Radish           & 95.65 & 95.65 & 95.65  & 95.00 & 100.00 & 100.00 \\
& Sweet\_Potato    & 47.62 & 52.38 & 52.38  & 65.00 & 100.00 & 100.00 \\
& Tomato           & 23.81 & 33.33 & 38.10  & 40.00 & 100.00 & 100.00 \\
\bottomrule
\end{tabular}
}
\vspace{0.5em}
\caption{Ingredient-wise cross-modal retrieval performance.}
\label{tab:suppl_ingredient_retrieval}
\end{table*}
\clearpage
\begin{table*}[t]
\centering
\small
\setlength{\tabcolsep}{6pt}
\renewcommand{\arraystretch}{1.05}
\resizebox{0.55\columnwidth}{!}{
\begin{tabular}{l|l|cc}
\toprule
\textbf{Family} & \textbf{Ingredient} & \textbf{mAP} & \textbf{mIoU} \\
\midrule

\multirow{10}{*}{Fruits}
& Apple            & 0.9240 & 0.7381 \\
& Banana           & 0.8411 & 0.6325 \\
& Kiwi             & 0.8446 & 0.6286 \\
& Lemon            & 0.8532 & 0.6589 \\
& Mandarin\_Orange  & 0.8289 & 0.6126 \\
& Mango            & 0.8625 & 0.6705 \\
& Peach            & 0.9283 & 0.7612 \\
& Pear             & 0.9024 & 0.7060 \\
& Pineapple        & 0.8518 & 0.6429 \\
& Strawberry       & 0.8806 & 0.6686 \\
\midrule

\multirow{10}{*}{Herbs}
& Angelica         & 0.9192 & 0.7485 \\
& Chamomile        & 0.5330 & 0.3594 \\
& Chives           & 0.7520 & 0.5320 \\
& Coriander        & 0.8941 & 0.7058 \\
& Dill             & 0.8009 & 0.5794 \\
& Garlic           & 0.5720 & 0.4002 \\
& Mint             & 0.5046 & 0.3814 \\
& Mugwort          & 0.9071 & 0.7192 \\
& Oregano          & 0.8057 & 0.6072 \\
& Turnip           & 0.6738 & 0.4893 \\
\midrule

\multirow{10}{*}{Nuts}
& Almond           & 0.8116 & 0.5494 \\
& Brazil\_Nut       & 0.9005 & 0.7293 \\
& Cashew           & 0.7745 & 0.6001 \\
& Chestnuts        & 0.9358 & 0.7607 \\
& Hazelnut         & 0.6982 & 0.4933 \\
& Peanuts          & 0.8733 & 0.6973 \\
& Pecans           & 0.8591 & 0.6827 \\
& Pili\_Nut         & 0.7117 & 0.5233 \\
& Pistachios       & 0.9619 & 0.8288 \\
& Walnuts          & 0.7324 & 0.5589 \\
\midrule

\multirow{10}{*}{Spices}
& Allspice         & 0.9040 & 0.7129 \\
& Chervil          & 0.8169 & 0.6068 \\
& Cinnamon         & 0.8618 & 0.6506 \\
& Cloves           & 0.9539 & 0.7764 \\
& Cumin            & 0.9371 & 0.7406 \\
& Ginger           & 0.8519 & 0.6352 \\
& Mustard          & 0.8858 & 0.6971 \\
& Nutmeg           & 0.8963 & 0.6929 \\
& Saffron          & 0.8945 & 0.6729 \\
& Star\_Anise       & 0.8126 & 0.6097 \\
\midrule

\multirow{10}{*}{Vegetables}
& Asparagus        & 0.9005 & 0.7174 \\
& Avocado          & 0.8448 & 0.6616 \\
& Broccoli         & 0.9145 & 0.7279 \\
& Brussel\_Sprouts  & 0.7275 & 0.5270 \\
& Cabbage          & 0.9268 & 0.7479 \\
& Cauliflower      & 0.7869 & 0.5770 \\
& Potato           & 0.9528 & 0.8074 \\
& Radish           & 0.8761 & 0.6890 \\
& Sweet\_Potato     & 0.9310 & 0.7537 \\
& Tomato           & 0.7970 & 0.6077 \\
\bottomrule
\end{tabular}
}
\vspace{0.5em}
\caption{Ingredient-wise smell localization performance.}
\label{tab:suppl_ingredient_localization}
\end{table*}
\clearpage

\section{Window Size Ablation}\label{sec:ablation}
\begin{table}[h]
\vspace{-8mm}
\centering
\resizebox{\textwidth}{!}{
\small
\setlength{\tabcolsep}{7pt}
\begin{tabular}{lcc|ccc|ccc|cc}
\toprule
\multirow{3.5}{*}{\textbf{Model}} &
\multicolumn{2}{c|}{\textbf{Smell Classification}} &
\multicolumn{6}{c|}{\textbf{Cross-Modal Retrieval}} &
\multicolumn{2}{c}{\textbf{Smell Localization}} \\

&  &  &
\multicolumn{3}{c|}{\textbf{Smell $\rightarrow$ Vision}} &
\multicolumn{3}{c|}{\textbf{Vision $\rightarrow$ Smell}} &
 &  \\
\cmidrule(r){4-6}\cmidrule(l){7-9}

&\textbf{Acc.} & \textbf{F1} &
\textbf{R@1} & \textbf{R@5} & \textbf{R@10} &
\textbf{R@1} & \textbf{R@5} & \textbf{R@10} & \textbf{mAP}
& \textbf{mIoU} \\
\midrule

\rowcolor{gray!15}
\multicolumn{11}{l}{$W=30$} \\
Global-CLIP & 50.30 & 49.37 & \underline{49.86} & 52.74 & 54.42 & \underline{46.00} & \textbf{83.10} & \textbf{93.00} & 0.1615 & 0.2072 \\
Ours-Global & 50.30 & 49.57 & 49.32 & 53.55 & 55.56 & 37.50 & 74.80 & 83.00 & 0.2752 & 0.2511 \\
Ours-Local  & \underline{51.82} & \underline{51.28} & 49.76 & \textbf{55.83} & \textbf{58.70} & 45.70 & 78.70 & 84.80 & \underline{0.7781} & \underline{0.5908} \\
\rowcolor{azure!10}
See \& Sniff & \textbf{52.20} & \textbf{51.31} & \textbf{50.14} & \underline{54.42} & \underline{57.07} & \textbf{51.30} & \textbf{83.10} & \underline{89.90} & \textbf{0.8002} & \textbf{0.6066} \\
\midrule

\rowcolor{gray!15}
\multicolumn{11}{l}{$W=50$} \\
Global-CLIP & 53.19 & 52.22 & 53.28 & 55.96 & 58.82 & \underline{56.90} & \textbf{87.40} & \underline{91.40} & 0.1736 & 0.2073 \\
Ours-Global & 53.74 & 52.64 & 50.97 & \textbf{62.60} & \textbf{67.50} & 52.80 & 78.30 & 85.30 & 0.6676 & 0.5016 \\
Ours-Local  & \underline{54.94} & \underline{53.78} & \underline{53.65} & 58.45 & 61.50 & 55.40 & 82.90 & 88.90 & \underline{0.7970} & \underline{0.6099} \\
\rowcolor{azure!10}
See \& Sniff & \textbf{57.71} & \textbf{56.68} & \textbf{56.14} & \underline{60.94} & \underline{63.90} & \textbf{63.20} & \textbf{87.40} & \textbf{91.90} & \textbf{0.8362} & \textbf{0.6456} \\
\midrule

\rowcolor{gray!15}
\multicolumn{11}{l}{$W=100$} \\
Global-CLIP & 61.55 & 60.01 & \textbf{61.95} & 63.35 & 64.94 & 58.00 & 82.30 & 92.00 & 0.1722 & 0.2029 \\
Ours-Global & 59.36 & 57.96 & 59.56 & 64.34 & 66.33 & 57.50 & 81.40 & 87.20 & 0.6438 & 0.4806 \\
Ours-Local  & \underline{62.75} & \underline{62.31} & 59.96 & \underline{67.53} & \underline{71.31} & \underline{62.70} & \underline{87.80} & \textbf{93.70} & \underline{0.8499} & \underline{0.6627} \\
\rowcolor{azure!10}
See \& Sniff & \textbf{63.75} & \textbf{62.66} & \underline{60.76} & \textbf{68.53} & \textbf{72.31} & \textbf{64.70} & \textbf{88.90} & \underline{93.30} & \textbf{0.8527} & \textbf{0.6654} \\
\bottomrule
\end{tabular}
}
\vspace{0.5em}
\caption{\textbf{Extended Results with $W=30$, $W=50$ and $W=100$.}}
\label{tab:suppl_extendedresult}
\vspace{-8mm}
\end{table}

\noindent In~\Tref{tab:odor_cls} of the main paper (Smell Classification Task), we also report results with $W{=}100$, following~\cite{feng2025smellnet}, while our default setup uses $W{=}50$. In this section, we report extended results for all downstream tasks across window sizes of 30, 50, and 100 to see the impact of window size. As shown in~\Tref{tab:suppl_extendedresult}, increasing the window size consistently improves performance across all metrics, indicating that longer temporal contexts help capture richer olfactory signatures. However, because our olfactory signals are derived from fixed 10-minute recordings, larger windows substantially reduce the number of available sniff units, lowering data granularity and overall data volume. Specifically, increasing $W$ from 30 to 50 and 100 reduces the training set from 9,265 to 5,411 and 2,512 samples, and the test set from 1,845 to 1,083 and 502, respectively. Beyond data scale, shorter windows are also better aligned with biological olfaction, which relies on brief sniffing rather than prolonged exposure, and they improve the practical viability of future systems that must operate on short olfactory signals for data efficiency. Therefore, we adopt $W{=}50$ as the default setting to balance performance, data granularity, and biological alignment with brief sniffing patterns.

\section{Visual Backbone Ablation}\label{sec:visbackbone}
\begin{table}[h]
\centering
\resizebox{0.7\linewidth}{!}{%
\begin{tabular}{lcccc}
\toprule
\textbf{Method} & \multicolumn{2}{c}{\textbf{Smell Classification}} & \multicolumn{2}{c}{\textbf{Smell Localization}} \\
\cmidrule(lr){2-3} \cmidrule(lr){4-5}
 & \textbf{Acc.} & \textbf{F1} & \textbf{mAP} & \textbf{mIoU} \\
\midrule
\rowcolor{cyan!10}
See\&Sniff (DINOv3-S)$\bigstar$     & 57.71 & 56.68 & 0.8362 & 0.6456 \\
\rowcolor{cyan!10}
See\&Sniff (DINOv3-B)     & \underline{58.91} & \underline{57.85} & \underline{0.8551} & \underline{0.6675} \\
\rowcolor{cyan!10}
See\&Sniff (DINOv3-L)     & \textbf{59.93} & \textbf{58.70} & \textbf{0.8593} & \textbf{0.6712} \\
Global-CLIP (CLIP-L)      & 53.19 & 52.22 & 0.1736 & 0.2073 \\
Global-SigLIP (SigLIP-L)  & 55.22 & 54.16 & 0.2072 & 0.2126 \\
See\&Sniff (CLIP-L)       & 53.83 & 53.60 & 0.4109 & 0.3046 \\
See\&Sniff (SigLIP-L)     & 57.89 & 56.30 & 0.5695 & 0.3962 \\
\bottomrule
\end{tabular}%
}
\vspace{2mm}
\caption{Ablation Study on the Visual Backbone.}
\vspace{-2em}
\label{tab:reb_vis_ablation}
\end{table}
\noindent We test various visual backbones and report results on Smell Classification (unimodal) and Localization (cross-modal) in~\Tref{tab:reb_vis_ablation}. As DINOv3 scales up, performance improves consistently as expected. We also evaluate CLIP and SigLIP~\cite{zhai2023sigmoid} backbones under global and local alignment settings. Global variants perform reasonably on Smell Classification but barely on Localization, as they do not exploit spatial information during training. Applying dense local alignment substantially improves localization, showing that it strengthens spatial grounding regardless of the backbone. However, a large gap to DINO remains, likely because global image-text contrastive objectives do not explicitly encourage local feature learning for dense prediction, as noted in ~\cite{naeem2024silc}. This analysis validates our pipeline design.

\section{Comparison with Feature Distillation Objectives}\label{sec:featdistll}
\textbf{Comparison with Feature Distillation Objectives.} We compare contrastive learning with feature distillation (FD) under both global and local alignment settings. \textit{Overall, our \textit{See \& Sniff} (contrastive learning) consistently outperforms FD across all metrics}. Global FD distills the DINOv3 CLS token into the pooled smell embedding, while local FD directly mirrors our dense local alignment by replacing max-similarity patch selection with min-distance patch selection, $\mathcal{L}_{\text{FD-local}}=\min_{h,w} dist(\bar{f}_o, f_v[h,w])$. We report the results with cosine distance in~\Tref{tab:reb_FD}.
While FD-Global achieves comparable classification and better localization than Ours-Global (likely because direct distillation preserves compatibility with DINO spatial features), contrastive learning performs better for retrieval, where fine-grained cross-modal discrimination is critical. FD-Local with both aligners collapses, as the learnable vision aligner acts as an unstable teacher during distillation. Removing the vision aligner (FD-Local$_{w/o VA}$) avoids collapse by directly distilling from a fixed teacher, but still underperforms \textit{See \& Sniff} and even FD-Global. This suggests that positive-only supervision can be unstable and less discriminative, as the positive pair is not guaranteed to be the true odor source and is not contrasted against competing regions or samples. In contrast, \textit{See \& Sniff} incorporates contrastive negatives, enforcing discriminative alignment and guiding the model toward odor-relevant regions.
\begin{table}[h]
\centering
\label{tab:feature_distillation_results}
\resizebox{0.85\linewidth}{!}{
\begin{tabular}{lccccc}
\toprule
\textbf{Method} & \textbf{Objective}
& \textbf{Cls. Acc} 
& \textbf{S$\rightarrow$V R@1}
& \textbf{V$\rightarrow$S R@1}
& \textbf{Loc. mIoU} \\
\midrule
\rowcolor{cyan!10}
See\&Sniff$\bigstar$ & Contrastive
& \textbf{57.71}
& \textbf{56.14}
& \textbf{63.20}
& \textbf{0.6456} \\
\rowcolor{cyan!10}
Ours-Global & Contrastive
& 53.74
& \underline{50.97}
& \underline{52.80}
& 0.5016 \\
\midrule

FD-Global & Distillation
& \underline{53.83}
& 49.22
& 40.50
& \underline{0.5666} \\

FD-Local$_{Both}$ & Distillation
& 2.95
& 2.22
& 2.20
& 0.2171 \\

FD-Local$_{w/oVA}$ & Distillation
& 53.37
& 46.70
& 43.10
& 0.5100 \\
\bottomrule
\end{tabular}
}
\vspace{2mm}
\caption{Comparison of Contrastive and Distillation Objectives.}
\vspace{-4mm}
\label{tab:reb_FD}
\end{table}

\section{Unseen Zero-shot Experiments}\label{sec:zeroshot}
\begin{figure}[h]
  \centering
  \begin{minipage}[t]{0.40\linewidth}
    \vspace{1.5pt}
    \centering
    \small
    \resizebox{0.85\linewidth}{!}{%
        \begin{tabular}{lcc}
        \toprule
        \textbf{Method} & \textbf{Acc.} & \textbf{F1} \\
        \midrule
        Chance      & 20.00 & - \\
        \midrule
        SmellNet      & 46.58 & 42.63 \\
        Global-CLIP      & \underline{51.14} & \underline{50.70} \\
        \rowcolor{cyan!10}
        See\&Sniff$\bigstar$     & \textbf{52.97} & \textbf{52.19} \\
        \bottomrule
        \end{tabular}%
    }
    \vspace{-1.5mm}
    \captionof{table}{Zero-shot Family-wise Classification.}
    \label{tab:reb_zeroshot_classification}
  \end{minipage}
  \hfill
  \begin{minipage}[t]{0.55\linewidth}
    \vspace{1.5pt}
    \centering
    \includegraphics[width=1\linewidth]{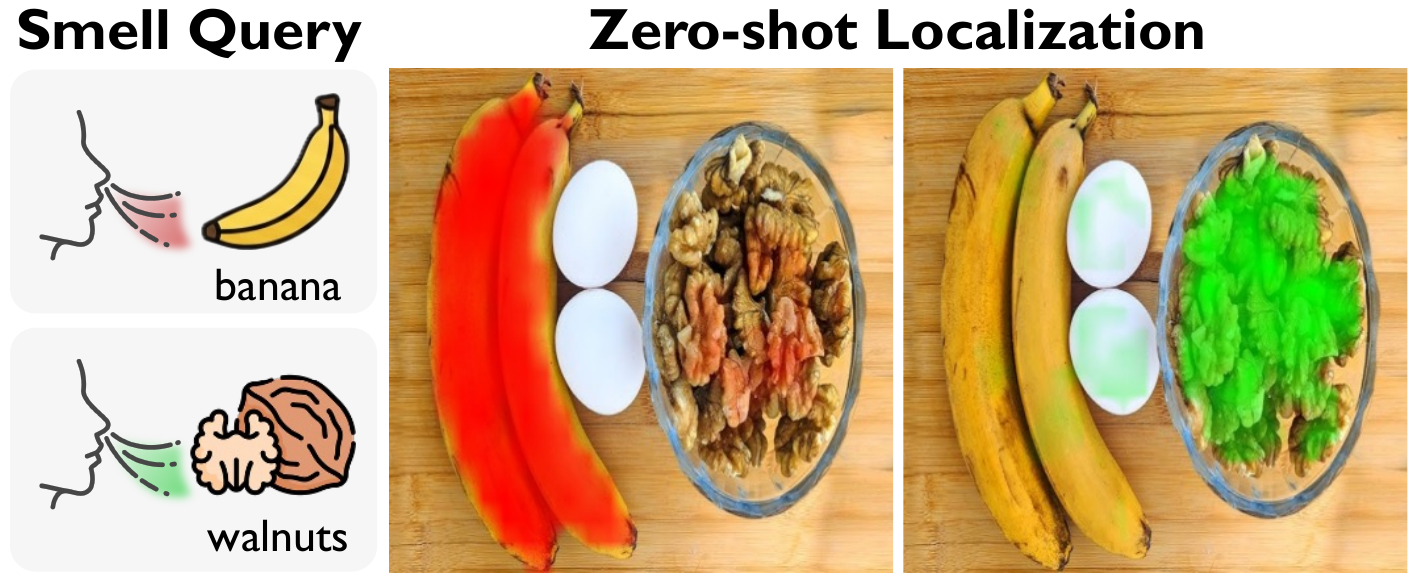}
    \vspace{-5.6mm}
    \captionof{figure}{Zero-shot Localization.}
    \label{fig:reb_zeroshot_localization}
  \end{minipage}
  \vspace{-4mm}
\end{figure}

Exact zero-shot ingredient classification is challenging in our current closed-set setting, since unseen categories have no learned classifier output or decision boundary. We therefore evaluate zero-shot transfer at the semantic-family level. Specifically, we train on 40 SmellNet-V ingredients and hold out 10 unseen ingredients, two per family: mango, banana, nutmeg, saffron, walnuts, brazil nut, coriander, oregano, brussel sprouts, and tomato. Given odors from these held-out categories, we evaluate whether the prediction falls into the correct family. As shown in~\Tref{tab:reb_zeroshot_classification}, \textit{See \& Sniff} achieves the best performance. This suggests that with visuo-olfactory training and dense alignment, our model captures transferable \textit{coarse} semantics beyond seen ingredients to some extent. We also observe qualitative zero-shot localization capability in~\Fref{fig:reb_zeroshot_localization}, where the model localizes an unseen ingredient conditioned on its odor, indicating that it retains some ability to ground unseen odors. However, exact zero-shot remains an important future direction.

\vspace{-4mm}
\section{Additional Qualitative Results}\label{sec:quali}
\begin{figure*}[t!]
    \centering
    \includegraphics[width=0.98\linewidth]{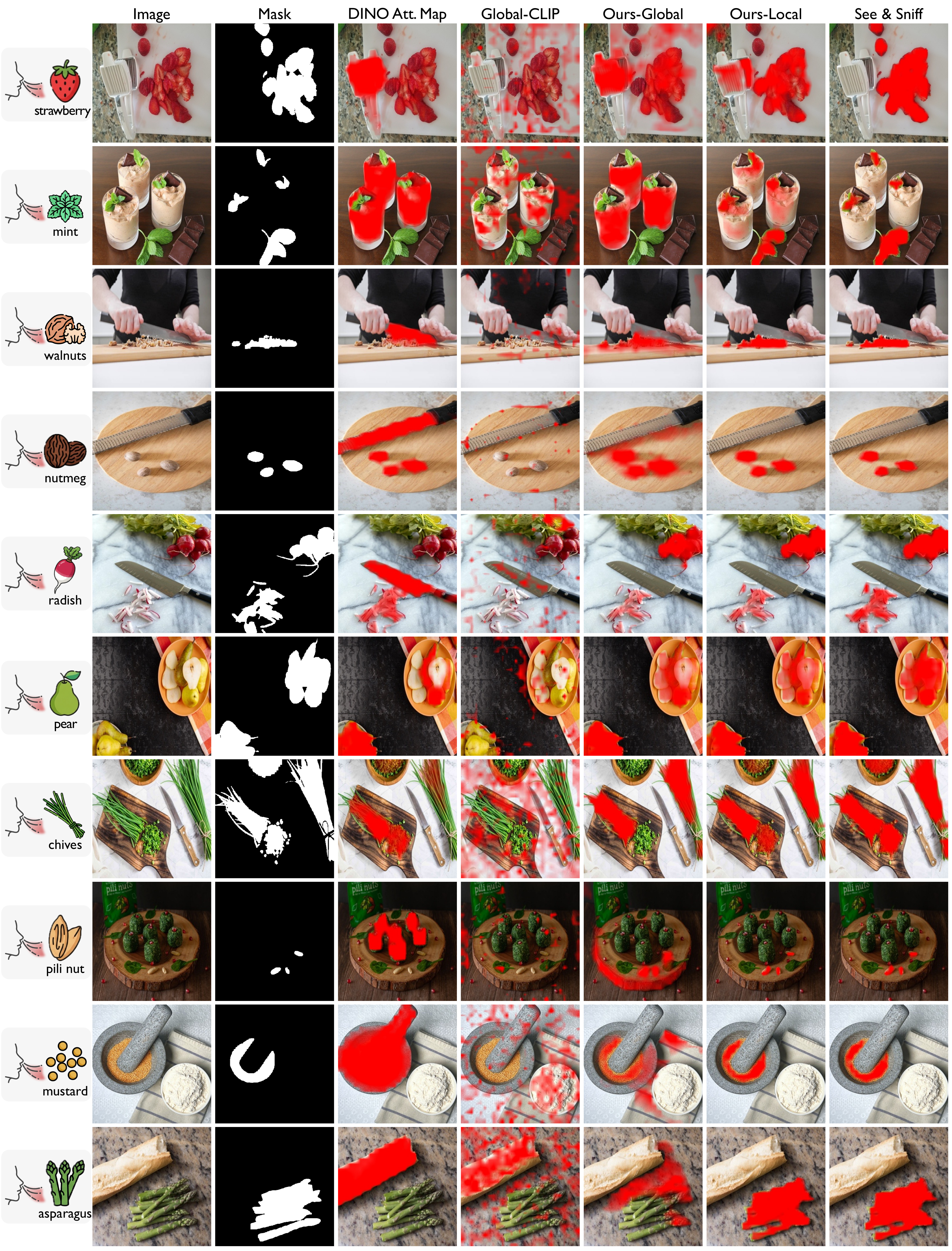}
    \vspace{-0.5em}
    \caption{\textbf{Qualitative Smell Localization Results.}}
    \label{fig:suppl_localization}
    \vspace{-2mm}
\end{figure*}

\begin{figure*}[t!]
    \centering
    \includegraphics[width=0.98\linewidth]{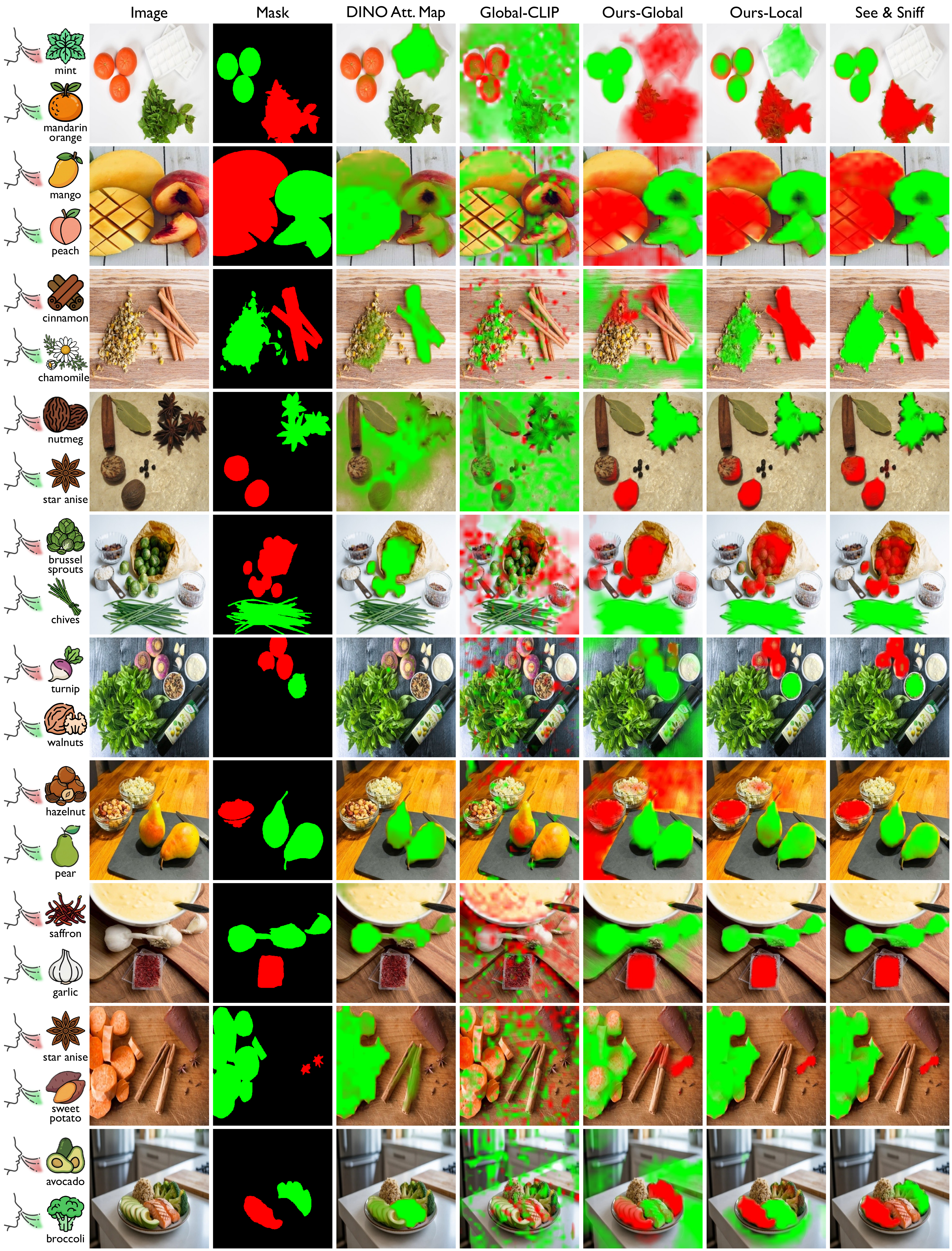}
    \vspace{-0.5em}
    \caption{\textbf{Qualitative Results on Interactive Localization.}}
    \label{fig:suppl_iiou}
    \vspace{-2mm}
\end{figure*}

Due to space constraints in the main paper, only a selected subset of qualitative results was presented. In this supplementary material, we provide comprehensive visualizations to further demonstrate the robustness of our model. ~\Fref{fig:suppl_localization} and ~\Fref{fig:suppl_iiou} present the results for smell localization and interactive localization, respectively. These visualizations demonstrate that \textit{See \& Sniff} can precisely distinguish and localize target ingredients based on specific smell queries.

\clearpage

\section{Comparison with Concurrent Unpublished Work}\label{sec:nyc}

As discussed in the main text, the concurrent work by Ozguroglu \etal~\cite{ozguroglu2025new} also investigates visuo-olfactory learning. While their study is dedicated to collecting a naturally synchronized visuo-olfactory dataset, we take a different approach by expanding an existing smell-only dataset via synthetic pairing with web images based on odor invariance.
This relatively straightforward approach for constructing visuo-olfactory data is efficient and effective, as demonstrated by our extensive experiments.
Architecturally, whereas their method focuses on establishing global contrastive alignment from a newly collected dataset of naturally paired signals, our approach is designed around dense local alignment using synthetic pairing based on odor invariance. This structural difference allows our method to not only learn global representations but also establish fine-grained spatial correspondences, thereby enabling novel downstream tasks such as visuo-olfactory grounding (i.e., smell localization).

Furthermore, performing a direct empirical comparison with Ozguroglu \etal is currently unfeasible. As an unpublished concurrent manuscript, its source code had not yet been publicly released at the time of our submission. Additionally, the paper lacks sufficient implementation details regarding their model architectures such as the specific layer configurations of their CNN or transformer olfaction encoders, and essential training hyperparameters. This omission poses significant challenges for exact reproduction. However, as mentioned in Section 4.2 - Baselines of the main paper, we approximate their architecture using our global baselines, as the methods are methodologically similar. Ultimately, despite these differences and current limitations in direct comparability, both works show the emerging importance and potential of advancing visuo-olfactory multimodal learning.

\end{document}